\documentclass{article}

    \PassOptionsToPackage{numbers, compress}{natbib}
 \usepackage[main, final]{neurips_2026}
\usepackage[utf8]{inputenc} % allow utf-8 input
\usepackage[T1]{fontenc}    % use 8-bit T1 fonts
\usepackage{graphicx}       % figures
\graphicspath{{./}{../}}    % shared pics/ folder lives one level up
\usepackage{hyperref}       % hyperlinks
\usepackage{subcaption}     % subfigures and subtables
\usepackage{url}            % simple URL typesetting
\usepackage{booktabs}       % professional-quality tables
\usepackage{tabularx}       % tables with equal-width flexible columns
\usepackage{amsfonts}       % blackboard math symbols
\usepackage{nicefrac}       % compact symbols for 1/2, etc.
\usepackage{microtype}      % microtypography
\usepackage{xcolor}         % colors
\usepackage[nointegrals]{wasysym}

\usepackage{amsmath,amssymb,amsthm}
\usepackage{thmtools}    % Optional but recommended for better formatting
\usepackage{algorithm}

\usepackage{listings}
\usepackage{xcolor}
\usepackage{courier}  
\usepackage{algpseudocode}
\usepackage{enumitem}
\newtheorem{definition}{Definition}[section]
\newtheorem{theorem}{Theorem}[section]
\newtheorem{proposition}{Proposition}[section]

\everypar{\looseness=-1}
\allowdisplaybreaks

\gdef\critic{critic}   

\newcommand{\new}[1]{{\color{black}#1}}

\title{Midpoint Generative Models}
\author{
    Daniil Shlenskii\textsuperscript{*,1,2}\\
    \And
    Nikita Gushchin\textsuperscript{*,2,1}\\
    \And
    Lev Novitskiy\textsuperscript{3,2}\\
    \AND
    Dmitry V. Dylov\textsuperscript{2,1}\\
    \And
    Alexander Korotin\textsuperscript{2,1}
}
\begin{document}

\footnotetext[1]{
    AXXX, Russia;
    \textsuperscript{2}Applied AI Institute, Russia;
    \textsuperscript{3}Kandinsky Lab, Russia
}
\begingroup
\renewcommand{\thefootnote}{\fnsymbol{footnote}}
\footnotetext[1]{
    Equal contribution.
    Correspondence to: \texttt{daniil.shlenskii@gmail.com}
}
\endgroup

\maketitle
\vspace{-15pt}
\begin{abstract}
We introduce \emph{Midpoint Generative Models} (MGM), a principled framework for training one-step generative models. MGM is based on a simple symmetry of Flow Matching with linear interpolation: when the two endpoint distributions coincide, the corresponding \new{velocity} field vanishes at the midpoint time, $t=1/2$. We show that the norm of this field defines a valid discrepancy between distributions, which we call the \emph{Midpoint Divergence}. We extend this discrepancy beyond the midpoint by introducing randomly flipped interpolations and further generalize it by replacing deterministic linear Flow Matching interpolations with symmetric stochastic interpolants, yielding a generalized Midpoint Divergence. Finally, we derive a variational formulation of our generalized divergence, yielding a tractable objective for training a one-step generator. The resulting MGM algorithm offers an effective and theoretically grounded approach to generative modeling, achieving competitive performance against existing one-step generative modeling methods.
\begin{center}
    \textbf{Code:} \url{https://github.com/daniil-shlenskii/mgm}
\end{center}
% \vspace{1mm}
\end{abstract}
\begin{figure}[htbp]
    \vspace{-15pt}
    \centering
    \includegraphics[width=1.0\textwidth]{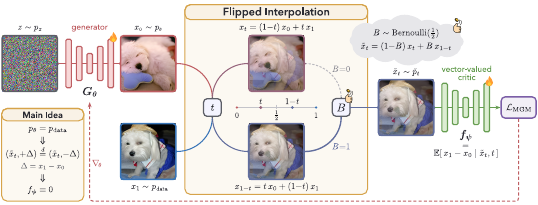}
    \caption{\textbf{Overview of our Midpoint Generative Models.}
        A latent $z \sim p_z$ is mapped by the generator $G_\theta$ to a sample $x_0\sim p_\theta$, which is then paired with a data sample $x_1 \sim p_{\mathrm{data}}$.
        The two endpoints are joined by a \emph{symmetric stochastic interpolant}~\eqref{eq:stoch_interpolant} at time $t \sim \mathcal{U}[0,1/2]$,
        producing both the interpolation $x_t \sim p_t$ and its \emph{time-flipped} counterpart $x_{1-t} \sim p_{1-t}$.
        A hidden flip $B$ then selects between the two, $\tilde{x}_t = (1-B)\,x_t + B\,x_{1-t}$, yielding the \emph{symmetrized observation} $\tilde{x}_t \sim \tilde{p}_t$  \new{(the flip $B$ itself is never observed)}.
        This flip is the key ingredient of MGM: by restoring endpoint symmetry, it forces the symmetrized displacement field $\tilde v_t$ in~\eqref{eq:vtb} \new{to vanish whenever $p_\theta = p_{\mathrm{data}}$}, making the generalized Midpoint Divergence~\eqref{eq:generalized_midpoint_divergence} a definite discrepancy (Theorem~\ref{thm:generalized_midpoint_divergence}).
        The \new{vector-valued \critic} $f_\psi$ approximates $\tilde{v}_t$ by regressing onto $x_1 - x_0$ given $\tilde{x}_t$ and $t$, while the generator $G_\theta$ minimizes the resulting divergence; the two are trained jointly in a minimax fashion through the variational midpoint objective~\eqref{eq:mgm_objective}.
        The complete procedure is detailed in Algorithm~\ref{alg:mgm_training_algo}.
    }
    \vspace{-5pt}
    \label{fig:teaser}
\end{figure}

\vspace{-2mm}
\section{Introduction}
\vspace{-2mm}

Diffusion and flow models are now at the core of many generative systems, enabling a wide range of
practical applications: image generation
\citep{dhariwal2021diffusion, rombach2022high}, video generation
\citep{ho2022video}, audio generation \citep{kong2021diffwave}, image editing
\citep{meng2022sdedit}, and image-to-image translation \citep{saharia2022palette}. However, whenever diffusion models are applied, the next question is
often the same: how can we make them fast enough for practical use? Several lines of work address
this question, including the development of more accurate numerical solvers
\citep{song2020denoising, karras2022elucidating, lu2022dpm}, the distillation of pre-trained
diffusion or flow models into one- or few-step generators
\citep{salimans2022progressive, song2023consistency, yin2024one, sauer2024adversarial,
gu2023boot}, and the design of new training schemes that learn one-step generators from scratch
using structural properties of diffusion and flow models
\citep{song2023consistency, song2024improved, frans2025one, geng2025mean}.

In this work, we propose a novel and effective framework for one-step generative modeling, built on a symmetry of Flow Matching with linear interpolation \citep{lipman2022flow, liu2022flow}. The starting observation is simple: under linear interpolation between identical endpoint distributions $p_0 = p_1$, the velocity field vanishes at the midpoint time $t = 1/2$:
\[
    v_{1/2}(x) \equiv 0 .
\]
% Conversely, a nonzero midpoint field witnesses a difference between the endpoints, which suggests measuring discrepancy through this field. We make this precise by defining a divergence as the expected squared norm of the midpoint velocity, which vanishes exactly when the two distributions coincide. This equivalence was also established by \citet{wang2026zero} as their \emph{zero-flow criterion} and used for representation learning and certifying conditional independence; we instead develop it into a divergence and a one-step generative objective.
\new{
    The converse holds as well: if $v_{1/2}\equiv 0$, then $p_0=p_1$.
    This equivalence was recently established by \citet{wang2026zero} as the \emph{zero-flow criterion} and
    used to learn sufficient representations through a squared-field loss.
    Building on this criterion, we define the \emph{Midpoint Divergence}
    as the expected squared norm of the midpoint velocity: a nonnegative
    discrepancy that vanishes if and only if $p_0=p_1$ and serves as a
    distribution-matching objective for training one-step generators.
}

\new{The vanishing property, however, is specific to the midpoint $t = 1/2$}: at any other time, the velocity field $v_t$ does not vanish even when $p_0 = p_1$, and therefore cannot certify equality of the endpoints on its own.
\new{
    To overcome this, we introduce a random time-flip $t \mapsto 1-t$ in the interpolation and show
    that the resulting symmetrized field vanishes at every $t$ whenever $p_0=p_1$, which extends the
    construction to a time-integrated Midpoint Divergence.
}
Moreover, the same construction applies beyond deterministic linear paths and yields a generalized Midpoint Divergence for symmetric
stochastic interpolants \citep{albergo2023building, albergo2025stochastic}.
This divergence is the basis of our training algorithm, Midpoint Generative Models (MGM),
illustrated in Figure~\ref{fig:teaser}.

\textbf{Our main contributions are as follows:}
\begin{itemize}[leftmargin=*, align=left]
    \vspace{-1.5mm}
    \item We introduce the \emph{Midpoint Divergence}, a new distributional discrepancy derived from the midpoint symmetry of linear interpolant Flow Matching (\wasyparagraph\ref{subsec:midpoint_divergence}).
    We prove that this divergence is definite: it is nonnegative and vanishes if and only if the two endpoint distributions are equal.
    \new{By introducing a random time flip, we further extend this construction to all interpolation times (\wasyparagraph\ref{subsec:time_midpoint_divergence}) and to symmetric stochastic interpolants (\wasyparagraph\ref{subsec:generalized_midpoint_divergence}), preserving definiteness in both cases.}

    \item We develop \emph{Midpoint Generative Models} (MGM), a practical training framework for one-step generative modeling based on the proposed divergence.
    We derive a variational formulation (\wasyparagraph\ref{subsec:midpoint_generative_models}) with a learned \new{vector-valued \critic}, yielding an effective algorithm for training generators directly from data,
    \new{without any teacher-model training signal}.
    % We show that our method is competitive with the other approaches, learning one-step generators based on diffusion/flow properties (\wasyparagraph\ref{sec:experiments}).
    We show that our \new{MGM is competitive with existing one-step generators on unconditional CIFAR-10, class-conditional ImageNet $64\times64$ in pixel space, and class-conditional ImageNet $512\times512$ in latent space} (\wasyparagraph\ref{sec:experiments}).
\end{itemize}

\vspace{-3mm}
\section{Background}
\vspace{-2mm}

\new{
    In this section, we briefly review Flow Matching (\wasyparagraph\ref{subsec:flow_matching}) and stochastic interpolants (\wasyparagraph\ref{subsec:stoch_interpolants}), which provide the velocity field and the interpolation paths underlying our construction.
}

\vspace{-2mm}
\subsection{Flow Matching}\label{subsec:flow_matching}
\vspace{-2mm}

Flow Matching~\citep{lipman2022flow,liu2022flow} learns a time-dependent \new{velocity} field that defines an ordinary differential
equation (ODE) transporting samples between distributions. Given independent endpoint samples
$X_0\sim p_0$ and $X_1\sim p_1$, a basic choice is the linear interpolant
\[
    X_t=(1-t)X_0+tX_1, \qquad t\in[0,1].
\]
This interpolant defines a path of marginal distributions $p_t=\mathrm{Law}(X_t)$.
These marginals are recovered by the probability flow ODE
\[
    \frac{dY_t}{dt}=v_t(Y_t), \qquad Y_0\sim p_0,
\]
where the \new{velocity field} is given by the conditional expectation
\begin{gather}\label{eq:vt}
    v_t(x)
    =
    \mathbb{E}_{X_0,X_1}
    \left[
        X_1-X_0 \mid X_t=x
    \right].
\end{gather}
Thus, Flow Matching reduces generative modeling to learning the \new{velocity field} $v_t(x)$
that transports samples along the marginal path from $p_0$ to $p_1$ through the ODE.

\vspace{-2mm}
\subsection{Stochastic Interpolants}
\label{subsec:stoch_interpolants}
\vspace{-2mm}

The stochastic interpolants framework is a generalization of the linear one, allowing more flexible, possibly noisy paths between endpoint samples \citep{albergo2025stochastic}. Let
$X_0\sim p_0$, $X_1\sim p_1$, and $\epsilon\sim\mathcal{N}(0,\mathrm{Id})$ be independent. A
stochastic interpolant is defined as
\begin{align*}
    X_t = I_t(X_0,X_1)+\sigma_t\epsilon,
    \qquad t\in[0,1],
\end{align*}
where
\[
    I_0(X_0,X_1)=X_0,\qquad
    I_1(X_0,X_1)=X_1,\qquad
    \sigma_0=\sigma_1=0,
\]
with $I_t$ twice continuously differentiable in both time and space and
$\sigma_t^2\in C^2([0,1])$.

The standard Flow Matching construction is recovered by setting
$I_t(X_0,X_1)=(1-t)X_0+tX_1$ and $\sigma_t=0$, giving the linear interpolant above.

\vspace{-2mm}
\section{Midpoint Generative Models}\label{sec:method}
\vspace{-2mm}

In this section, we introduce \textbf{M}idpoint \textbf{G}enerative \textbf{M}odels (MGM), a principled framework for training one-step generative models.
We begin in \S\ref{subsec:midpoint_divergence} by showing that the velocity field induced by linear Flow Matching contains a discriminative signal at the midpoint $t=1/2$. In particular, this velocity field vanishes when the endpoint distributions coincide, and its failure to vanish gives rise to the Midpoint Divergence.
Next, in \S\ref{subsec:time_midpoint_divergence}, we extend this idea beyond the midpoint. Since non-midpoint times introduce an orientation bias, we introduce a randomly flipped interpolation that restores symmetry and leads to a time-integrated Midpoint Divergence.
In \S\ref{subsec:generalized_midpoint_divergence}, we further generalize the construction from deterministic linear interpolations to symmetric stochastic interpolants, yielding a generalized Midpoint Divergence.
Finally, in \S\ref{subsec:midpoint_generative_models}, we show how this generalized divergence can be used as a training objective for generative modeling. We derive a variational formulation with a learned \critic\ model and summarize the resulting practical training algorithm for one-step generators.
All \underline{proofs} are provided in Appendix~\ref{app:appendix}.

\vspace{-2mm}
\subsection{Midpoint Divergence}\label{subsec:midpoint_divergence}
\begin{figure}[t]
    \vspace{-3mm}
    \centering
    \makebox[\linewidth][c]{%
        \includegraphics[width=1.1\linewidth]{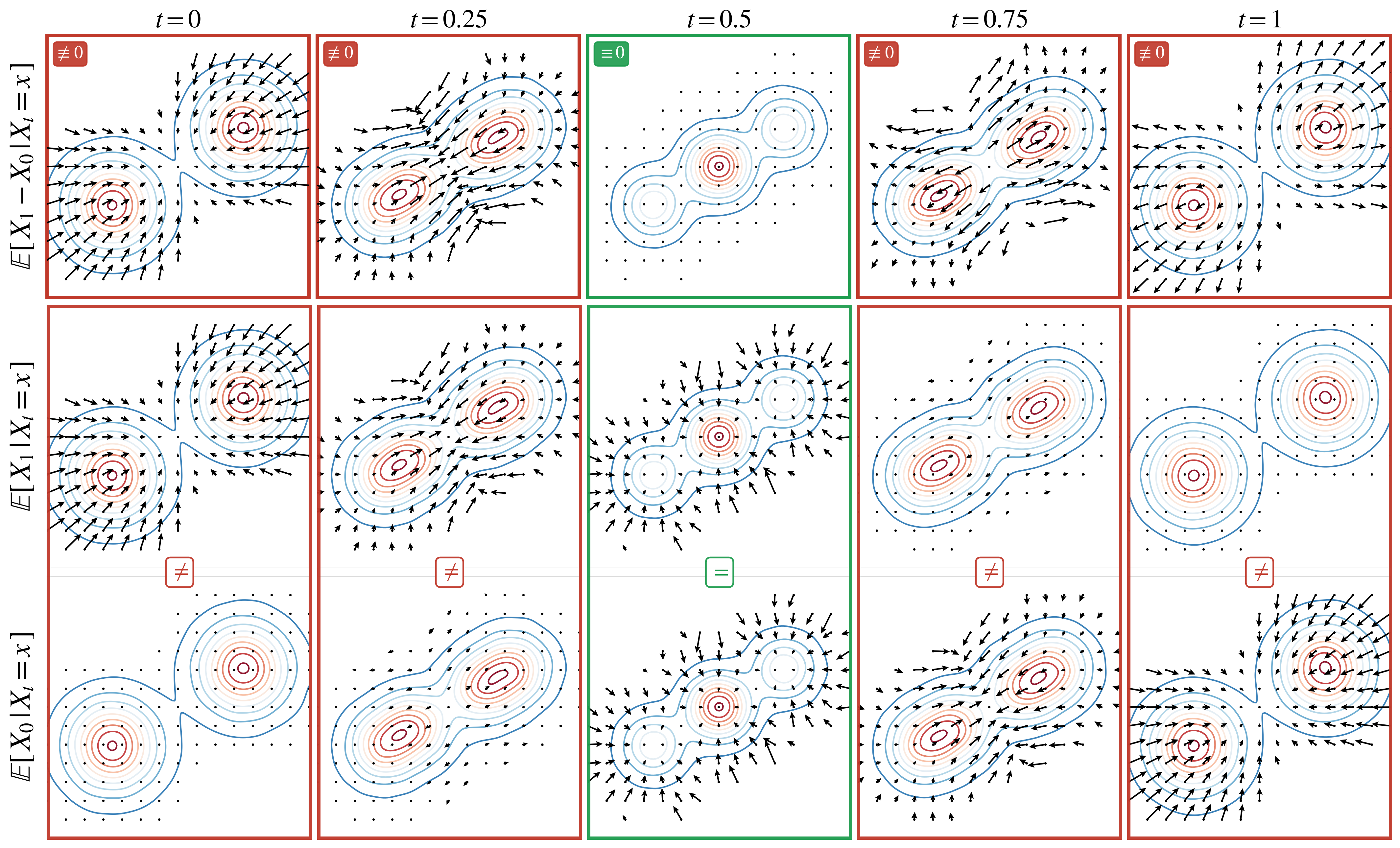}%
    }
    \caption{Linear interpolation for a Gaussian toy example. Contours show the interpolated density $p_t$.
    \emph{Top row:} The velocity field
    $v_t(x)=\mathbb{E}[X_1-X_0\mid X_t=x]$ vanishes only at $t=1/2$ and is nonzero away from the midpoint.
    \emph{Bottom rows:} the endpoint denoisers are related by time reversal,
    $\mathbb{E}[X_0 \mid X_t=x] = \mathbb{E}[X_1 \mid X_{1-t}=x]$, rather than by equality at a fixed time.}
    \label{fig:midpoint-straight}
    \vspace{-4mm}
\end{figure}

\vspace{-2mm}
Consider the Flow Matching linear interpolation between two distributions $p_0$ and $p_1$
and the corresponding velocity field $v_t$ in~\eqref{eq:vt}.
\new{A key property of this field is that it vanishes at the midpoint $t=1/2$ whenever $p_0=p_1$ (see Appendix~\ref{app:electrostatics} for an electrostatic interpretation).}
We illustrate this in Figure~\ref{fig:midpoint-straight} and state it formally below.

\begin{proposition}[Midpoint symmetry]\label{prop:midpoint_symmetry}
    Let $X_0$ and $X_1$ be independent samples from the same distribution $p$ with finite first moment.
    Then
    \begin{align*}
        v_{1/2}(x)
        =
        \mathbb{E}_{X_0,X_1}\!\left[X_1-X_0 \mid X_{1/2}=x\right]
        = 0
        \quad \text{for } p_{1/2}\text{-a.e. }x.
    \end{align*}
\end{proposition}

This suggests using the failure of the velocity field to vanish at the midpoint as a measure of discrepancy between $p_0$ and $p_1$.

\begin{definition}[Midpoint Divergence]\label{def:midpoint_divegence}
    For distributions $p_0$ and $p_1$, define the Midpoint Divergence
    \begin{align}\label{eq:midpoint_divegence}
        D_{\mathrm{mid}}(p_0, p_1)
        :=
        \mathbb{E}_{X_{1/2}}
        \left\|
            \mathbb{E}_{X_0,X_1}\!\left[X_1-X_0 \mid X_{1/2}\right]
        \right\|_2^2.
    \end{align}
\end{definition}

By Proposition~\ref{prop:midpoint_symmetry}, we immediately have
$D_{\mathrm{mid}}(p_0,p_1)=0$ whenever $p_0=p_1$.
To justify the term ``divergence,'' we now show the converse: the Midpoint Divergence vanishes only when the endpoint distributions coincide.

\begin{theorem}[Definiteness of the Midpoint Divergence]
    \label{thm:definiteness_of_midpoint_divergence}
    Let $X_0 \sim p_0$ and $X_1 \sim p_1$ be independent with sub-exponential tails. Then
    \[
    D_{\mathrm{mid}}(p_0,p_1)=0
    \quad\Longleftrightarrow\quad
    p_0=p_1.
    \]
\end{theorem}

Thus, vanishing of the midpoint Flow Matching velocity field characterizes equality of the endpoint distributions.
A natural next question is whether analogous discrepancies can be constructed using non-midpoint times.

\vspace{-2mm}
\subsection{Time-integrated Midpoint Divergence}\label{subsec:time_midpoint_divergence}
\vspace{-2mm}

\begin{figure}[t]
    \vspace{-3mm}
    \centering
    \makebox[\linewidth][c]{%
        \includegraphics[width=1.1\linewidth]{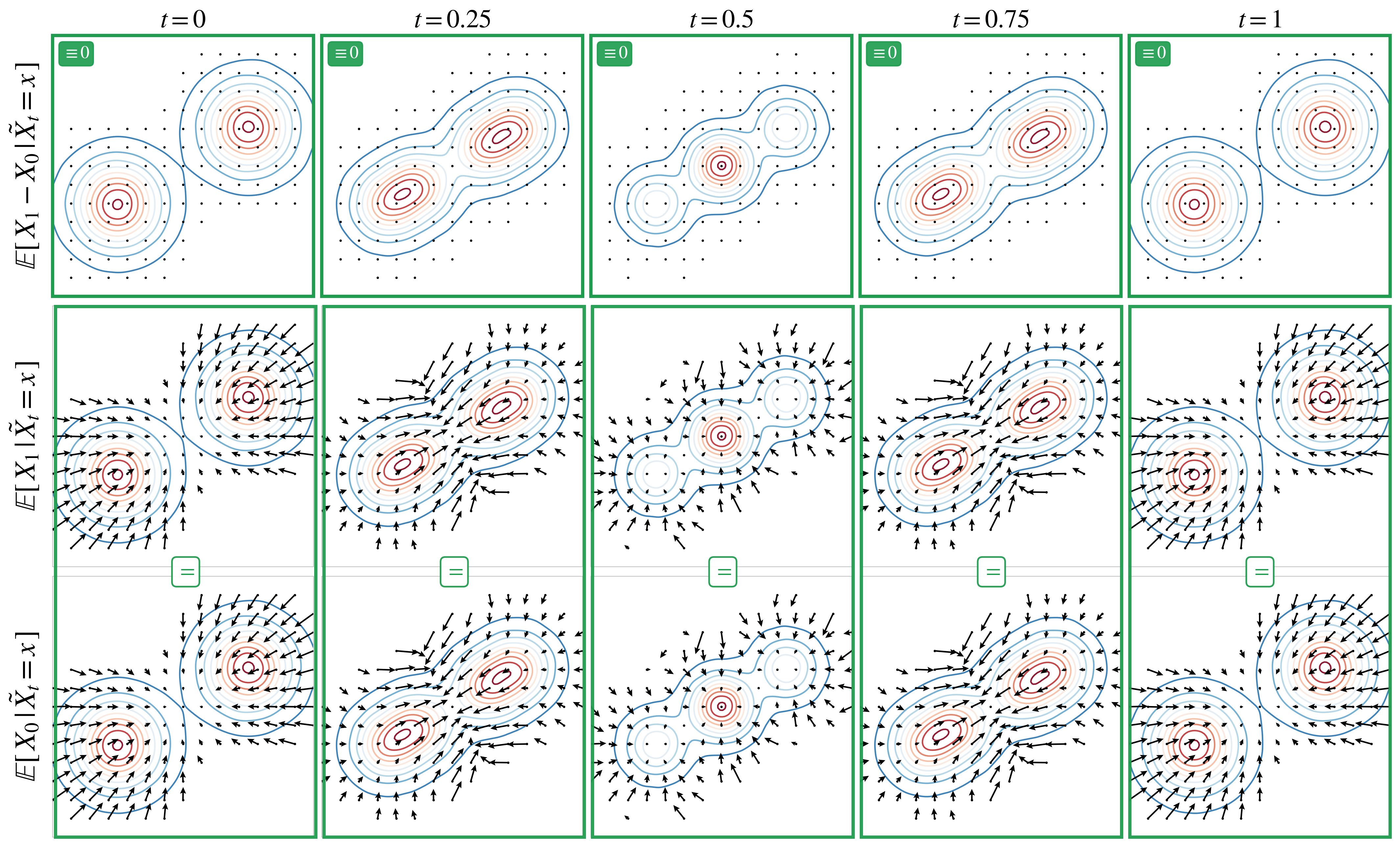}%
    }
    \caption{Randomly flipped interpolation on the same Gaussian toy example.
    Contours show the symmetrized density $\tilde{p}_t$.
    \emph{Top row:} The symmetrized
    displacement field $\tilde{v}_t(x)=\mathbb{E}[X_1-X_0\mid \tilde{X}_t=x]$ is
    identically zero for every displayed $t$.
    \emph{Bottom rows:} Under the same symmetrized
    observation model, the endpoint denoisers coincide pointwise for every $t$,
    $\mathbb{E}[X_0 \mid \tilde{X}_t=x] = \mathbb{E}[X_1 \mid \tilde{X}_t=x]$.}
    \label{fig:midpoint-flipped}
    \vspace{-4mm}
\end{figure}

A naive way to extend the midpoint construction over the full Flow Matching path is to integrate the squared norm of the Flow Matching velocity field over time:
\begin{align}\label{eq:naive_time_integrated_midpoint_divergence}
    D_{\mathrm{t\text{-}mid}}^{\mathrm{naive}}(p_0,p_1)
    :=
    \int_0^{1} \mathbb{E}_{X_t}
        \left\|
            \mathbb{E}_{X_0,X_1}[X_1-X_0 \mid X_t]
        \right\|_2^2 dt .
\end{align}
However, this naive functional is not a valid discrepancy: for $t\neq 1/2$, the Flow Matching velocity field $v_t$ generally does not vanish even when $p_0=p_1$.
Figure~\ref{fig:midpoint-straight} (top row) illustrates this failure mode in a Gaussian toy example: under linear interpolation, the velocity field vanishes at the midpoint but remains nonzero at the other displayed times.
This can be seen \new{more clearly} by rewriting the velocity as
a difference of two posterior endpoint means, which we refer to as \emph{denoisers}:
\begin{align*}
    v_t(x)
    =
    \mathbb{E}_{X_0,X_1}\left[X_1-X_0 \mid X_t=x\right]
    =
    \mathbb{E}_{X_1}\left[X_1\mid X_t=x\right]
    -
    \mathbb{E}_{X_0}\left[X_0\mid X_t=x\right].
\end{align*}
Even when $p_0=p_1$, these two denoisers are generally different away from the
midpoint. For example, at $t=0$ we have $X_t=X_0$, and hence
\begin{align*}
    \mathbb{E}_{X_1}\left[X_1\mid X_0=x\right]
    =
    \mathbb{E}_{X_1}X_1,
    \qquad
    \mathbb{E}_{X_0}\left[X_0\mid X_0=x\right]
    =
    x.
\end{align*}
Thus, given $X_0=x$, the squared-error optimal estimator of $X_0$ is $x$ itself, whereas
the optimal reconstruction of the independent endpoint $X_1$ is only its mean.
The same phenomenon occurs at other non-midpoint times: the closer $t$ is to one
endpoint, the easier it is to reconstruct that endpoint from $X_t$, and the harder
it is to reconstruct the other endpoint.
Hence, the two denoisers are intrinsically
asymmetric away from $t=1/2$.
This phenomenon is visually illustrated in Figure~\ref{fig:midpoint-straight} (bottom rows).
To remove this orientation bias, we introduce a \emph{randomly flipped interpolation}
\begin{align}\label{eq:flip_flow_matching_interpolant}
    \widetilde{X}_t =
    \begin{cases}
        X_t = (1-t)X_0 + tX_1, & B=0,\\
        X_{1-t} = t X_0 + (1-t)X_1, & B=1,
    \end{cases}
\end{align}
\new{where $B\sim \mathrm{Bernoulli}(1/2)$ is independent of $(X_0,X_1)$.}
Importantly, the flip variable $B$ is not observed.
This randomness makes the two endpoints equally close, in
distribution, to the observed interpolation point.
Consequently, when $p_0=p_1$, neither endpoint has an intrinsic
reconstruction advantage, and \new{the expected displacement given $\widetilde{X}_t$} vanishes.
This is illustrated in Figure~\ref{fig:midpoint-flipped} and formalized in the following proposition.
\begin{proposition}[Flip-induced symmetry]\label{prop:flip_symmetry}
    Let $X_0$ and $X_1$ be independent samples from the same distribution $p$ with
    finite first moment. Then, for every $t\in[0,1]$,
    the symmetrized displacement field
    \begin{align}\label{eq:vtb}
        \widetilde{v}_t(x)
        :=
        \mathbb{E}_{X_0,X_1}
        [
            X_1-X_0 \mid \widetilde{X}_t = x
        ]
        =
        0
        \quad \text{for } \widetilde{p}_{t}\text{-a.e. }x,
    \end{align}
    where $\widetilde{X}_t \sim \widetilde{p}_{t} := (p_t + p_{1-t})/2$.
\end{proposition}
\new{At $t=1/2$, both branches coincide, so $\widetilde v_{1/2}=v_{1/2}$ and the flipped construction reduces to the midpoint one.}
Since, for $p_0=p_1$, the symmetrized displacement field vanishes at every $t$,
it is natural to define a discrepancy by integrating its squared norm over time.

\begin{definition}[Time-integrated Midpoint Divergence]\label{def:time_integrated_midpoint_divergence}
    For any two distributions $p_0$ and $p_1$ we define time-integrated Midpoint Divergence as the following functional
    \begin{align*}
        D_{\mathrm{t\text{-}mid}}(p_0,p_1) := \int_0^{1/2} \mathbb{E}_{\widetilde{X}_t}
            \left\|
                \mathbb{E}_{X_0,X_1}[X_1-X_0 \mid \widetilde{X}_t]
            \right\|_2^2 dt\new{.}
    \end{align*}
\end{definition}
We integrate only over $[0,1/2]$ because, under the random flip construction, the discrepancy at time $t$ is the same as at time $1-t$;
integrating over $[0,1]$ would therefore count each contribution twice.
As in the single-time midpoint case, this functional is a valid divergence.

\begin{theorem}[Definiteness of the time-integrated Midpoint Divergence]\label{thm:time_intergrated_midpoint_divergence}
    Let $X_0 \sim p_0$ and $X_1 \sim p_1$ be independent with finite second moments. Then
    \[
        D_{\mathrm{t\text{-}mid}}(p_0,p_1)=0
        \quad\Longleftrightarrow\quad
        p_0=p_1.
    \]
\end{theorem}
% Figure~\ref{fig:midpoint-flipped} shows the same toy example after random
% time flip: the symmetrized displacement field is zero for every $t$,
% and the two endpoint denoisers coincide pointwise.

\vspace{-2mm}
\subsection{Generalized Midpoint Divergence}\label{subsec:generalized_midpoint_divergence}
\vspace{-2mm}
In the flipped construction, the symmetrized displacement  field $\widetilde{v}_t(x)$ in~\eqref{eq:vtb}
is no longer the Flow Matching velocity field in~\eqref{eq:vt}.
Nevertheless,
it can still be interpreted as the difference between two posterior endpoint
denoisers under the symmetrized observation model:
\begin{align*}
    \widetilde{v}_t(x)
    =
    \mathbb E_{X_0,X_1}[X_1 - X_0 \mid \widetilde{X}_t = x]
    =
    \mathbb E_{X_1}[X_1 \mid \widetilde{X}_t = x]
    -
    \mathbb E_{X_0}[X_0 \mid \widetilde{X}_t = x].
\end{align*}
This perspective decouples the construction from the specific linear
interpolation used in Flow Matching and suggests a broader class of symmetric
interpolation paths.

We now extend the construction to stochastic interpolants defined in \S\ref{subsec:stoch_interpolants}:
\begin{align}\label{eq:stoch_interpolant}
    X_t = I_t(X_0, X_1) + \sigma_t \epsilon,
    \qquad
    \epsilon \sim \mathcal N(0, \mathrm{Id}).
\end{align}
We additionally require the interpolant to be symmetric, namely
\[
I_t(x_0,x_1)
=
I_{1-t}(x_1,x_0),
\qquad
\sigma_t=\sigma_{1-t}.
\]

Let $B \sim \mathrm{Bernoulli}(1/2)$ be independent of $(X_0, X_1, \epsilon)$.
We define the flipped observation by
\begin{align}\label{eq:flipped_stochastic_interpolant}
    \widetilde{X}_t =
    \begin{cases}
        X_t = I_t(X_0, X_1) + \sigma_t \epsilon, & B = 0,\\
        X_{1-t} = I_{1-t}(X_0, X_1) + \sigma_{1-t} \epsilon, & B = 1.
    \end{cases}
\end{align}
This symmetrized observation model leads to the following generalized Midpoint Divergence:
\begin{align}\label{eq:generalized_midpoint_divergence}
    D^{I, \sigma}_{\mathrm{t\text{-}mid}}(p_0, p_1)
    :=
    \int_0^{1/2}
    \mathbb E_{\widetilde{X}_t}
    \left\|
        \mathbb E_{X_0,X_1}[X_1 - X_0 \mid \widetilde{X}_t]
    \right\|_2^2
    \, dt.
\end{align}
This reduces to the time-integrated midpoint divergence for the Flow Matching linear interpolant.
The next theorem shows that this generalized construction still defines a valid divergence.

\begin{theorem}[Definiteness of the generalized Midpoint Divergence]\label{thm:generalized_midpoint_divergence}
    Let $X_0 \sim p_0$ and $X_1 \sim p_1$ be independent with finite second moments.
    Fix any symmetric stochastic interpolant~\eqref{eq:stoch_interpolant}.
    Then
    \[
        D^{I, \sigma}_{\mathrm{t\text{-}mid}}(p_0, p_1) = 0
        \quad\Longleftrightarrow\quad
        p_0 = p_1.
    \]
\end{theorem}
Theorem~\ref{thm:generalized_midpoint_divergence} shows that the generalized Midpoint Divergence can serve as a principled training objective: minimizing it against the data distribution identifies the target distribution uniquely. We now use this divergence to construct our Midpoint Generative Models.

\vspace{-2mm}
\subsection{Midpoint Generative Models}\label{subsec:midpoint_generative_models}
\vspace{-2mm}
The generalized Midpoint Divergence provides a natural approach for generative modeling.
Let
$p_\theta$ be the distribution induced by a generator $G_\theta$ applied to a
latent prior $p_z$:
\[
    Z \sim p_z,
    \qquad
    X_0 = G_\theta(Z),
    \qquad
    X_0 \sim p_\theta .
\]
We pair generated samples with data samples
$X_1 \sim p_{\rm data}$, and define $\widetilde X_t$ using the flipped
stochastic interpolant in~\eqref{eq:flipped_stochastic_interpolant}.
The generalized midpoint divergence then provides a natural training objective
\begin{align}\label{eq:mgm_primal_objective}
    \min_\theta
    D^{I, \sigma}_{\mathrm{t\text{-}mid}}(p_\theta, p_{\rm data})
    =
    \int_0^{1/2}
    \mathbb E_{\widetilde{X}_t}
    \left\|
        \mathbb E_{X_0,X_1}[X_1 - X_0 \mid \widetilde{X}_t]
    \right\|_2^2
    \, dt.
\end{align}
By Theorem~\ref{thm:generalized_midpoint_divergence}, the divergence vanishes
if and only if $p_\theta=p_{\rm data}$.
Directly optimizing~\eqref{eq:mgm_primal_objective}, however, is intractable
because the \new{inner expectation} is unknown. We therefore
use the following variational representation.

\begin{proposition}[Variational midpoint objective]\label{prop:variation_midpoint_objective}
For fixed $p_\theta$, the generalized midpoint divergence admits the variational
form
\[
D^{I,\sigma}_{\mathrm{t\text{-}mid}}
(p_\theta,p_{\rm data})
=
\max_{f_t}
\int_0^{1/2}
\mathbb E_{X_0, X_1, B, \epsilon}
\left[
2\langle f_t(\widetilde X_t), X_1-X_0\rangle
-
\|f_t(\widetilde X_t)\|_2^2
\right]
dt,
\]
where $\widetilde{X}_t$ is defined according to~\eqref{eq:flipped_stochastic_interpolant} and the maximum is over square-integrable vector-valued functions
$f_t$. Moreover, the maximizer is given by
\[
    f_t^\star(x)
    =
    \mathbb E_{X_0,X_1}\left[
        X_1-X_0
        \mid
        \widetilde X_t=x
    \right]
    =
    \widetilde{v}_t(x),
\]
for almost every $t$ and $\widetilde p_t$-almost every $x$.
\end{proposition}

Using Proposition~\ref{prop:variation_midpoint_objective} and parameterizing the symmetrized displacement field by a neural network $f_\psi$,
\new{which we call the vector-valued \emph{\critic},}
we obtain the following minimax objective:
\begin{align}\label{eq:mgm_objective}
    \min_\theta \max_\psi \;
    \mathcal L_\mathrm{MGM}(\theta,\psi)
    :=
    \int_0^{1/2}
    \mathbb E_{X_0, X_1, B, \epsilon}
    \left[
    2\langle f_\psi(\widetilde X_t,t), X_1-X_0\rangle
    -
    \|f_\psi(\widetilde X_t,t)\|_2^2
    \right]dt .
\end{align}
\new{The integrand of~\eqref{eq:mgm_objective} can be equivalently rewritten as}
\[
    \new{
    2\langle f_\psi(\widetilde X_t,t), X_1-X_0\rangle
    -
    \|f_\psi(\widetilde X_t,t)\|_2^2
    =
    \|X_1-X_0\|_2^2 - \|f_\psi(\widetilde X_t,t) - (X_1-X_0)\|_2^2 .}
\]
\new{Since the first term does not depend on $\psi$, the \critic\ update is exactly Flow Matching regression of $X_1-X_0$, applied to the symmetrized observation $\widetilde X_t$ instead of $X_t$. The generator, in contrast, must use the original form~\eqref{eq:mgm_objective}, since $\|X_1-X_0\|_2^2$ depends on $\theta$ through $X_0=G_\theta(Z)$.}
In practice, we alternate between updating the \critic\ model to approximate the
symmetrized displacement field and updating the generator to minimize the resulting
variational midpoint objective. The full stochastic training procedure is summarized in
Algorithm~\ref{alg:mgm_training_algo}.

\textbf{Importance of flipping and time integration.}
The variational formulation above can be instantiated not only with the generalized Midpoint Divergence, but also with simpler alternatives such as the midpoint-only divergence
$D_{\mathrm{mid}}$ in~\eqref{eq:midpoint_divegence}
and the naive unflipped time-integrated objective
$D_{\mathrm{t\text{-}mid}}^{\mathrm{naive}}(p_0,p_1)$ in~\eqref{eq:naive_time_integrated_midpoint_divergence}; details are provided in Appendix~\ref{app:algorithm}.
These alternatives allow us to ablate the two key ingredients of MGM: random flipping and time integration.

\begin{figure}[t]
\vspace{-0mm}
    \centering
        \includegraphics[width=0.245\linewidth]{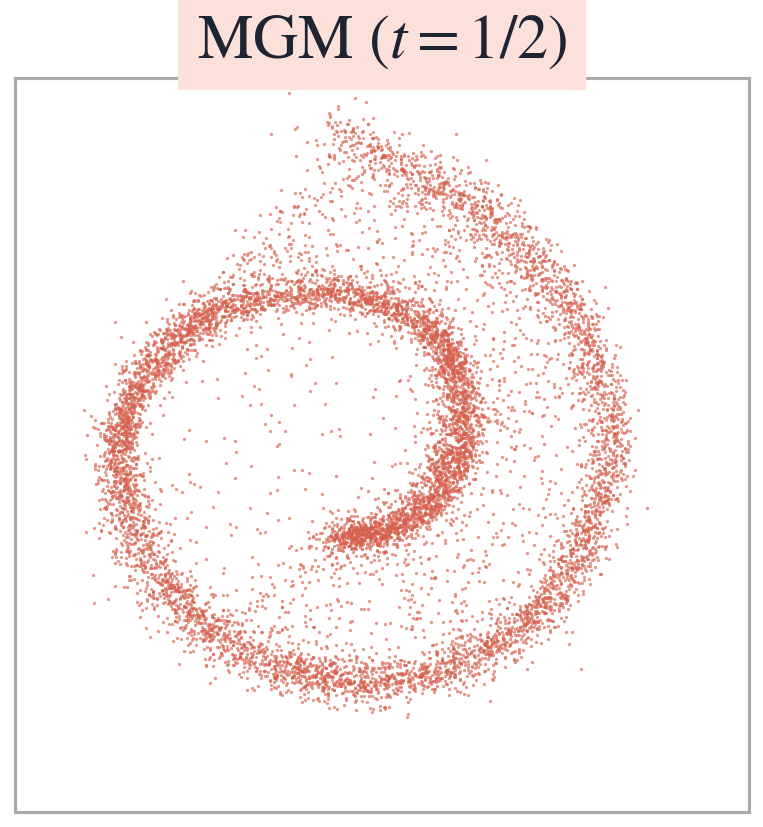}\hfill
        \includegraphics[width=0.245\linewidth]{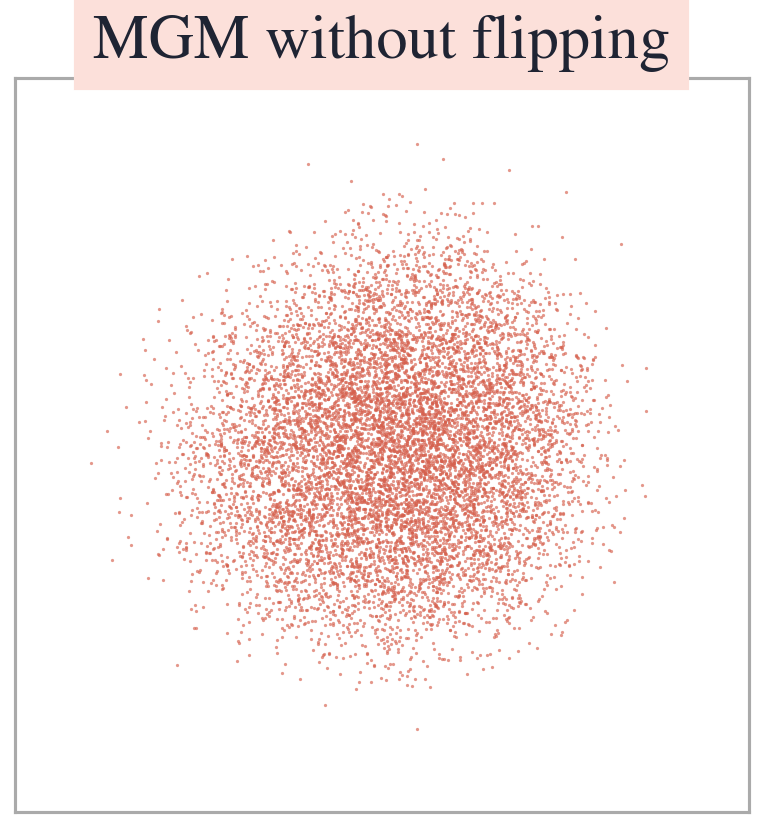}\hfill
        \includegraphics[width=0.245\linewidth]{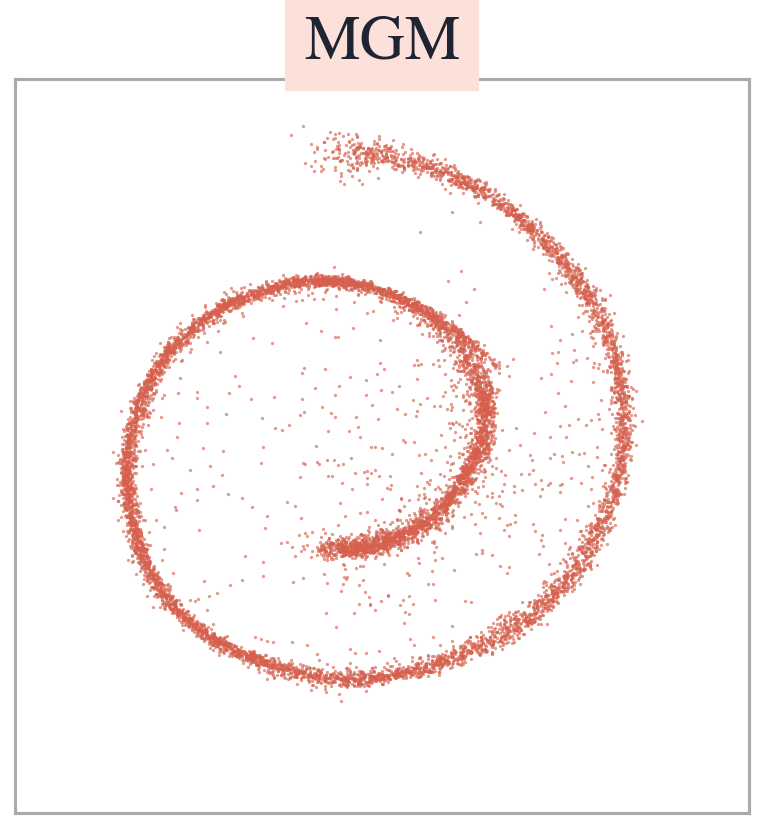}\hfill
        \includegraphics[width=0.245\linewidth]{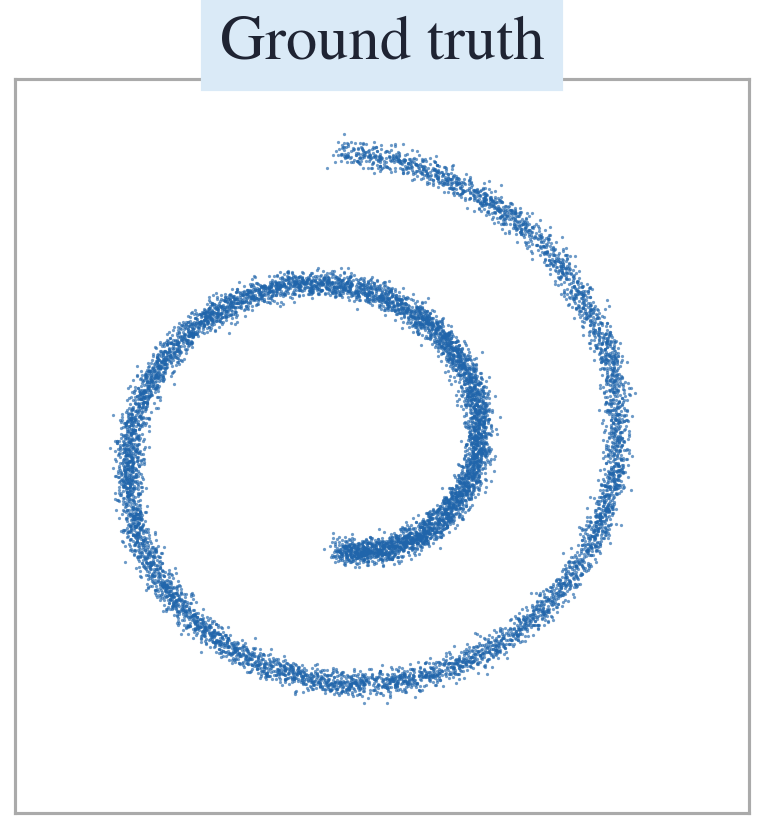}\hfill
    \caption{Toy example illustrating the role of the MGM construction.
    Full MGM recovers the target distribution, while the midpoint-only objective at $t=1/2$ remains valid but produces noisier samples.
    In contrast, naively integrating the unflipped midpoint objective over time fails.}
    \vspace{-4mm}
    \label{fig:toy_samples}
\end{figure}

Figure~\ref{fig:toy_samples} shows the resulting samples on the Swiss roll toy dataset.
The midpoint-only objective learns the coarse geometry of the target, but produces noisy samples.
One possible explanation is that, at $t=1/2$, the \critic\ $f_\psi$ only observes the mixed interpolation
$x_{1/2}=(x_0+x_1)/2$, in which the generator output $x_0$ appears only through an averaged state.
This may provide a weaker signal for correcting fine-grained errors in generator samples.
In contrast, the flipped time-integrated objective includes times closer to the endpoints.
For small $t$, the symmetrized observation contains samples close to either $X_0$ or $X_1$, while preserving endpoint symmetry through the random flip.
Thus, time integration gives access to a richer family of observations, including near-generator samples, which can provide a stronger signal for fine-grained details.

The naive unflipped time-integrated objective fails, as expected from the orientation bias of non-midpoint Flow Matching velocities.
The flipped time-integrated objective, however, accurately recovers the target distribution.
This comparison supports the full MGM construction: random flipping makes the symmetrized displacement field vanish when $p_0=p_1$, while time integration provides a stronger training signal than the midpoint alone.
\new{For the empirical benefit of the stochastic interpolants introduced in~\S\ref{subsec:generalized_midpoint_divergence}, see~\S\ref{subsec:ablation_study}.}

\vspace{-2mm}
\section{Related Work}
\vspace{-2mm}

Now we discuss three main axes of work related to MGM: distillation of pre-trained diffusion and flow models, direct training of fast generators from diffusion/flow principles, and GANs.

\textbf{Distillation from pre-trained diffusion and flow models.}
A common route to fast generation is to distill a pre-trained diffusion or flow model into a one- or few-step student. One branch exploits PF-ODE consistency: teacher samples follow deterministic trajectories, so a pair $(x_t,t)$ identifies the remaining trajectory including the final point. These methods train a student to reproduce two or more teacher-sampler steps or to make predictions at different points on the same teacher trajectory agree \citep{salimans2022progressive,song2023consistency,song2024improved,lu2025simplifying,kim2024consistency,berthelot2023tract,geng2025consistency,lee2025truncated}. A second branch parameterizes $p_\theta(x_0)$ through a stochastic generator and optimizes losses derived from a pre-trained teacher \citep{yin2024one,sauer2024adversarial,luo2023diffinstruct,yin2024improved,xu2025fdistill,zhou2024score,luo2024sim,gushchin2025inverse,huang2024flow,shlenskii2025overclocking}. These losses usually lead to adversarial optimization and often require an additional estimator for the generator distribution, such as a generator score, generator flow, or critic. Many of these works can be viewed through the inverse-distillation lens of Universal Inverse Distillation \citep{kornilov2025universal}. \textit{In contrast, MGM does not distill a pre-trained diffusion or flow teacher; it trains directly from data using the proposed Midpoint Divergence between generated samples and real data, without any teacher \new{signal in training loss}.}

\textbf{Training one- and few-step generators from scratch using diffusion/flow properties.}
Another line of work trains fast samplers directly from data, without a pretrained teacher as the training signal. These methods use the fact that finite-time updates along the same generative ODE trajectory must be mutually consistent: an update from $s$ to $t$ should agree with updates that pass through intermediate times. Their common goal is to obtain such finite-time integrators, parameterized as endpoint-consistent predictors, two-time trajectory maps, finite-interval updates, or average velocities \new{\citep{song2023consistency,song2024improved,lu2025simplifying,kim2024consistency,frans2025one,geng2025mean}}.
\new{
    MeanFlow~\citep{geng2025mean}, for example, learns an average velocity constrained to be consistent with the instantaneous velocity of the generative ODE, so the guidance toward the data is internal to the model itself.
    Some methods, such as Consistency Models and Consistency Trajectory Models, also support teacher distillation, but their direct-training variants use only data and noise samples \citep{song2023consistency,kim2024consistency}.
    Flow Map Matching gives a general view of many such objectives as learning maps between pairs of times in diffusion or stochastic-interpolant dynamics \citep{boffi2025flowmap}.
    MGM is close in spirit, as it also trains without a teacher using diffusion and flow principles.
    \textit{Unlike these approaches, MGM does not learn a finite-time integrator: its generator directly parameterizes the generated distribution, while the guidance toward the data comes from an auxiliary \critic\ discarded at inference.}
}

\textbf{GANs.}
Classical GANs \citep{goodfellow2014generative} train a generator against a scalar discriminator, with $f$-GAN \citep{nowozin2016fgan} and Wasserstein GAN \citep{arjovsky2017wasserstein} connecting discriminator-generator minimax objectives to standard divergences. Diffusion-GAN variants \citep{xiao2021denoising,wang2023diffusiongan} combine adversarial objectives with diffusion ideas, for example by discriminating noised samples or modeling large denoising steps adversarially. MGM also has a minimax training form, but uses a vector-valued critic $f_\psi(x,t)$ trained toward endpoint displacements under a symmetrized stochastic interpolant. \textit{Thus, MGM optimizes a regression-based generalized Midpoint Divergence, not a classification-based adversarial loss.}

% \begin{figure}[t]
%     \vspace{-3mm}
%     \centering
%     \begin{subfigure}[t]{0.48\textwidth}
%         \centering
%         \includegraphics[width=\linewidth,trim={384 416 384 416},clip]{pics/cifar_generations.png}
%         \caption{MGM}
%     \end{subfigure}\hfill
%     \begin{subfigure}[t]{0.48\textwidth}
%         \centering
%         \includegraphics[width=\linewidth,trim={384 416 384 416},clip]{pics/cifar_generations_t_1_2.png}
%         \caption{MGM, $t=1/2$ only}
%     \end{subfigure}
%     \vspace{-2mm}
%     \caption{Crops from uncurated CIFAR-10 sample grids generated with one network evaluation.
%     The time-integrated MGM objective preserves more fine-grained structure than the midpoint-only variant.}
%     \label{fig:cifar_generations}
%     \vspace{-5mm}
% \end{figure}

\makeatletter
\renewcommand{\p@subtable}{\thetable.}
\makeatother

\begin{table}[!t]
    \centering
    \normalsize
    \setlength{\tabcolsep}{2pt}
    \renewcommand{\arraystretch}{1.03}
    \captionsetup[subtable]{font=small,justification=centering}
    \makebox[\textwidth][c]{%
    \begin{minipage}{0.98\textwidth}
    \centering
    \begin{subtable}[t]{0.215\linewidth}
        \centering
        \begin{tabular*}{\linewidth}{@{\extracolsep{\fill}} c c @{}}
            \toprule
            Warmup & FID \\
            \midrule
            0K   & 34.02 \\
            10K  & 3.73 \\
            20K  & 2.69 \\
            \textbf{50K}  & \textbf{2.34} \\
            100K & 2.42 \\
            200K & 2.66 \\
            \bottomrule
        \end{tabular*}
        \caption{\textbf{Warmup duration}}
        \label{tab:ablation_warmup}
    \end{subtable}
    \hfill
    \begin{subtable}[t]{0.215\linewidth}
        \centering
        \begin{tabular*}{\linewidth}{@{\extracolsep{\fill}} c c @{}}
            \toprule
            $\sigma$ & FID \\
            \midrule
            0.00 & 5.87 \\
            0.01 & 2.58 \\
            \textbf{0.05} & \textbf{2.46} \\
            0.20 & 2.78 \\
            0.50 & 3.55 \\
            \phantom{0.50} & \phantom{3.55} \\
            \bottomrule
        \end{tabular*}
        \caption{\textbf{Interpolant stochasticity}}
        \label{tab:ablation_interpolant_stochasticity}
    \end{subtable}
    \hfill
    \begin{subtable}[t]{0.215\linewidth}
        \centering
        \begin{tabular*}{\linewidth}{@{\extracolsep{\fill}} c c @{}}
            \toprule
            Critic $\sigma$ & FID \\
            \midrule
            0.00 & 5.87 \\
            \textbf{0.01} & \textbf{2.27} \\
            0.05 & 2.34 \\
            0.20 & 2.98 \\
            0.50 & 56.58 \\
            \phantom{0.50} & \phantom{56.58} \\
            \bottomrule
        \end{tabular*}
        \caption{\textbf{Critic-only noise}}
        \label{tab:ablation_critic_stochasticity}
    \end{subtable}
    \hfill
    \begin{subtable}[t]{0.215\linewidth}
        \centering
        \begin{tabular*}{\linewidth}{@{\extracolsep{\fill}} c c @{}}
            \toprule
            Critic $\sigma$ & FID \\
            \midrule
            0.00 & 14.75 \\
            \textbf{0.01} & \textbf{13.55} \\
            0.05 & 19.76 \\
            0.20 & 68.62 \\
            0.50 & 120.48 \\
            \phantom{0.50} & \phantom{120.48} \\
            \bottomrule
        \end{tabular*}
        \caption{\textbf{Midpoint-only training}}
        \label{tab:ablation_midpoint_only}
    \end{subtable}
    \end{minipage}
    }
    \caption{\textbf{Ablation study on unconditional CIFAR-10 $32\times32$ generation.}
    FID-50K is reported for one-step generation, and the best setting within each ablation is in \textbf{bold}.
    % Performance improves up to 50K warmup steps and then saturates (\ref{tab:ablation_warmup}). A small amount of interpolant noise is beneficial, while excessive noise degrades FID (\ref{tab:ablation_interpolant_stochasticity}). The critic-only noise ablation keeps the generator interpolant deterministic but adds noise to the \critic\ observations; this gives the best result (\ref{tab:ablation_critic_stochasticity}). Midpoint-only training performs substantially worse than the time-integrated objective, confirming the benefit of sampling multiple interpolation times (\ref{tab:ablation_midpoint_only}).}
    }
    \label{tab:ablation_study}
    \vspace{-8mm}
\end{table}

\vspace{-2mm}
\section{Experiments}\label{sec:experiments}
\vspace{-2mm}

We evaluate MGM for one-step unconditional generation on CIFAR-10~\citep{krizhevsky2009learning} and class-conditional generation on ImageNet\new{~\citep{deng2009imagenet}}.
We use pixel-space models for CIFAR-10 and ImageNet $64\times64$, and latent-space models for ImageNet $512\times512$, following EDM2~\citep{karras2024analyzing}. We report FID~\citep{heusel2017gans} on 50K generated samples.
In \S\ref{subsec:experimental_setup}, we describe the stochastic interpolant and initialization used in our experiments.
In \S\ref{subsec:ablation_study}, we ablate the main design choices of MGM on CIFAR-10.
Finally, in \S\ref{subsec:baseline_comparison}, we compare MGM against recent one-step generative models.
Additional experimental details, further results, and computational cost analysis are provided in Appendices~\ref{app:experimental_details}, \ref{app:additional_results}, and~\ref{app:computational_cost}, respectively.

\vspace{-2mm}
\subsection{Experimental Setup}\label{subsec:experimental_setup}
\vspace{-2mm}
In all experiments, we train MGM with the generalized Midpoint Divergence (\S\ref{subsec:generalized_midpoint_divergence}), using the linear interpolant $I_t(x_0,x_1)=(1-t)x_0+tx_1$ with the noise schedule $\sigma_t=\sqrt{\sigma t(1-t)}$.
On ImageNet, both the generator and the \critic\ are initialized from pretrained diffusion models: EDM~\citep{karras2022elucidating} for $64\times64$ and EDM2-S~\citep{karras2024analyzing} for $512\times512$.
On CIFAR-10, we instead initialize from a short denoising warmup to show that MGM does not require a fully trained diffusion model.
In all cases, the pretrained network serves only as an initialization and does not enter the training loss.
The warmup is ablated in \S\ref{subsec:ablation_study} and detailed in Appendix~\ref{app:experimental_details}.

\begin{table}[t]
    \centering
    \setlength{\tabcolsep}{4pt}
    \renewcommand{\arraystretch}{1.03}
    % \ours: gray box with 1.1pt padding around digit height; fits the row strut, so rows do not grow
    \providecommand{\ours}[1]{{\setlength{\fboxsep}{0pt}\colorbox{blue!10}{\rule[-1.1pt]{0pt}{\dimexpr\fontcharht\font`0+2.2pt\relax}\hspace{2pt}#1\hspace{2pt}}}}
    % \second: underline placed 1pt below the baseline so that it stays inside \ours boxes
    \providecommand{\second}[1]{{\setbox0\hbox{#1}\rlap{\rule[-1pt]{\wd0}{0.4pt}}\box0}}
    \begin{tabular*}{\textwidth}{@{\extracolsep{\fill}}l c l c @{}}
        \toprule
        \multicolumn{2}{c}{w/ teacher-model training signal} &
        \multicolumn{2}{c}{w/o teacher-model training signal} \\
        \midrule
        Method & FID & Method & FID \\
        \midrule
        \multicolumn{4}{c}{\textit{Unconditional CIFAR-10 $32\times32$}} \\
        \midrule
        PD \citep{salimans2022progressive} & 8.34 & ECT \citep{geng2025consistency} & 3.60 \\
        2-Rectified Flow \citep{liu2022flow} & 4.85 & VCT \citep{silvestri2025vct} & 3.26 \\
        TRACT \citep{berthelot2023tract} & 3.78 & IMM \citep{zhou2025inductive} & 3.20 \\
        DFNO \citep{zheng2023fast} & 3.78 & StyleGAN2 w/ ADA \citep{karras2020training} & 2.92 \\
        DMD \citep{yin2024one} & 3.77 & Stable CT \citep{wang2025stable} & 2.92 \\
        sCD \citep{lu2025simplifying} & 3.66 & MeanFlow \citep{geng2025mean} & 2.92 \\
        CD \citep{song2023consistency} & 3.55 & sCT \citep{lu2025simplifying} & 2.85 \\
        CMT w/ ECT \citep{hu2026cmt} & 2.74 & iCT \citep{song2024improved} & 2.83 \\
        TCM \citep{lee2025truncated} & 2.46 & iCT-deep \citep{song2024improved} & 2.51 \\
        CTM \citep{kim2024consistency} & \second{1.98} & DiffRatio-DiJS \citep{chen2025diffratio} & \second{2.39} \\
        SiD \citep{zhou2024score} & \textbf{1.92} & \ours{MGM (Ours)} & \ours{\textbf{2.27}} \\
        \midrule
        \multicolumn{4}{c}{\textit{Class-conditional ImageNet $64\times64$ (pixel space)}} \\
        \midrule
        sCD \citep{lu2025simplifying} & 2.44 & VCT \citep{silvestri2025vct} & 4.93 \\
        ECD \citep{geng2025consistency} & 2.24 & iCT \citep{song2024improved} & 4.02 \\
        TCM \citep{lee2025truncated} & 2.20 & iCT-deep \citep{song2024improved} & 3.25 \\
        CMT w/ ECT \citep{hu2026cmt} & 2.02 & ECT \citep{geng2025consistency} & 2.49 \\
        CTM \citep{kim2024consistency} & 1.92 & Stable CT \citep{wang2025stable} & 2.42 \\
        CMT w/ ECD \citep{hu2026cmt} & 1.78 & sCT \citep{lu2025simplifying} & \second{2.04} \\
        SiD \citep{zhou2024score} & \second{1.52} & DiffRatio-DiJS \citep{chen2025diffratio} & \textbf{1.54} \\
        DMD2 \citep{yin2024improved} & \textbf{1.28} & \ours{MGM (Ours)} & \ours{2.50} \\
        \midrule
        \multicolumn{4}{c}{\textit{Class-conditional ImageNet $512\times512$ (latent space)}} \\
        \midrule
        ECD \citep{geng2025consistency} & 8.47 & sCT-S \citep{lu2025simplifying} & 10.13 \\
        CMT-XXL w/ ECD \citep{hu2026cmt} & 3.38 & ECT \citep{geng2025consistency} & 9.98 \\
        AYF-S \citep{sabour2026align} & 3.32 & sCT-XXL \citep{lu2025simplifying} & 4.29 \\
        sCD-S \citep{lu2025simplifying} & \second{3.07} & DiffRatio-DiJS-M \citep{chen2025diffratio} & \textbf{1.41} \\
        sCD-XXL \citep{lu2025simplifying} & \textbf{2.28} & \ours{MGM-S (Ours)} & \ours{\second{2.36}} \\
        \bottomrule
    \end{tabular*}
    \vspace{0.35em}
    \caption{One-step generation on unconditional CIFAR-10 and class-conditional ImageNet, grouped by whether a pretrained model enters the training loss (\S\ref{subsec:baseline_comparison}). Within each group, the best result is in \textbf{bold}, the second best is \underline{underlined}, and MGM is \ours{highlighted}. On ImageNet $512\times512$, suffixes denote EDM2 sizes: S (280M), M (498M), and XXL (1.5B). Baseline values are from the cited papers or as collected by CMT \citep{hu2026cmt}.}
    \label{tab:baseline_cifar10}
    \vspace{-3mm}
\end{table}

\vspace{-2mm}
\subsection{Ablation Study}\label{subsec:ablation_study}
\vspace{-2mm}
We ablate four design choices in MGM: how long to run the denoising-objective warmup, how much noise to use in the stochastic interpolant, whether the generator should observe noisy interpolants, and whether the objective should integrate over time or use only the midpoint. The corresponding FID results are reported in Table~\ref{tab:ablation_study}.

\textbf{Warmup duration.}
We vary the \new{denoising-objective warmup budget} before MGM training (Table~\ref{tab:ablation_warmup}).
\new{Following~\citet{karras2022elucidating}, we measure warmup budgets in kimgs (thousands of images processed during training).}
FID improves from 34.02 without warmup to 2.34 after 50K\new{~kimgs of warmup} (25\% of the original EDM training \new{budget}), with no further gains from longer warmups.
So we use the 50K\new{~kimgs} checkpoint to initialize all subsequent CIFAR-10 experiments.

\textbf{Interpolant stochasticity.}
We vary the noise level $\sigma$ in the stochastic schedule $\sigma_t=\sqrt{\sigma t(1-t)}$ (Table~\ref{tab:ablation_interpolant_stochasticity}).
Adding noise improves FID substantially over the deterministic setting $\sigma=0$, with small nonzero noise giving the best results.
This is consistent with observations in image-to-image translation that adding bridge noise can improve performance \citep{liu2022let,zhou2023denoising,zheng2024diffusion,he2024consistency,wang2025implicit}.
Increasing the noise further worsens FID.

\textbf{Critic stochasticity.}
We next decouple the generator and \critic\ interpolants (Table~\ref{tab:ablation_critic_stochasticity}) by adding noise only in the \critic\ loss, while retaining the random time flip in both losses.
For small noise levels, this critic-only stochasticity improves FID, giving the best result in our ablations, and for moderate levels it performs comparably to the fully stochastic setting. This suggests that noise is primarily important as a regularizer of the \critic.
\new{At the} large noise level $\sigma=0.5$, however, training degrades significantly, which may reflect the increased mismatch between the noisy points on which the \critic\ is trained and the noiseless points at which it is evaluated in the generator loss.

\textbf{Midpoint-only training.} We compare the time-integrated objective with a midpoint-only variant that fixes $t=1/2$ (Table~\ref{tab:ablation_midpoint_only}). In the same critic-only noise setting, the midpoint-only variant is substantially worse across all noise levels, reaching at best FID 13.55 compared to 2.27, and it breaks down already at moderate noise $\sigma=0.2$. This confirms that extending the divergence to all interpolation times via the random time flip (\S\ref{subsec:time_midpoint_divergence}) provides a much stronger learning signal than the midpoint alone.

\begin{figure}[t]
    \centering
    \includegraphics[width=\linewidth]{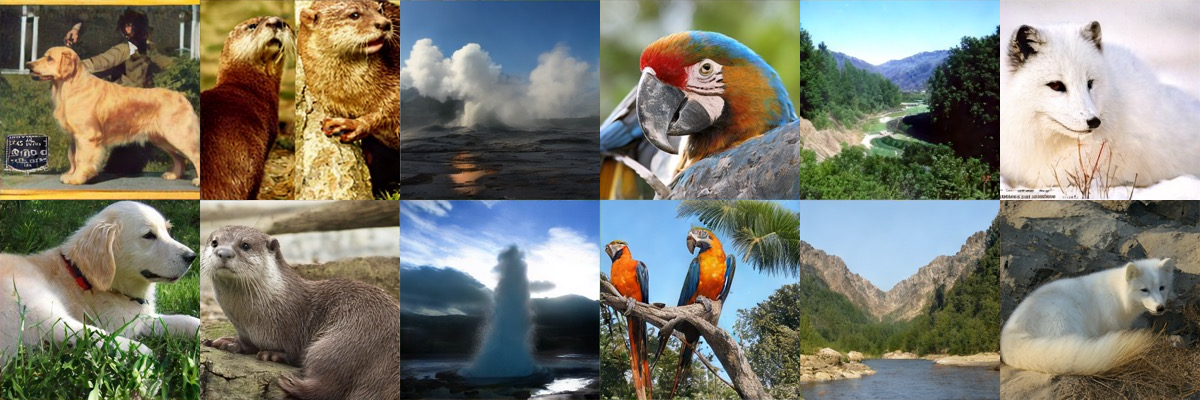}
    \caption{One-step ImageNet $512\times512$ samples from MGM (FID 2.36); more in Appendix~\ref{app:additional_samples}.}
    \label{fig:imagenet512_samples}
    \vspace{-3mm}
\end{figure}

\vspace{-2mm}
\subsection{Baseline Comparison}\label{subsec:baseline_comparison}
\vspace{-2mm}
We compare MGM with recent one-step generative models in Table~\ref{tab:baseline_cifar10}, grouping them by whether a teacher-model training signal is used, i.e., whether a separately pretrained model, or targets and trajectories produced by it, enters the generator loss; using a pretrained model only to initialize the networks does not count as such a signal.
On CIFAR-10, MGM achieves FID 2.27, the best result among methods without a teacher-model training signal and comparable to distillation-based approaches.
On ImageNet $64\times64$, MGM reaches FID 2.50, comparable to recent one-step methods.
On ImageNet $512\times512$, MGM with a 280M-parameter generator achieves FID 2.36, outperforming or approaching several 1.5B-parameter models, including teacher-based ones.
One-step ImageNet $512\times512$ samples are shown in Figure~\ref{fig:imagenet512_samples}.
Overall, MGM is comparable with recent one-step generative models across all benchmarks without relying on a teacher model during training.

\vspace{-2mm}
\section{Discussion and Limitations}\label{sec:discussion}
\vspace{-2mm}
We introduced \textit{Midpoint Generative Models} (MGM), a principled framework for training one-step generators by minimizing the generalized Midpoint Divergence.
As in GANs~\citep{goodfellow2014generative}, MGM updates the generator using a jointly trained \critic.
Unlike a GAN discriminator, however, the MGM \critic\ is vector-valued, like the auxiliary networks of distribution-matching distillation methods~\citep{yin2024one,luo2023diffinstruct,zhou2024score}.
In contrast to these methods, the \critic\ alone guides the generator, without a separate pretrained score network for the data distribution.
These properties, together with MGM's competitive results, make it a promising direction for one-step generative modeling.

At the same time, MGM has several limitations.
First, like other minimax methods, MGM can be unstable.
Second, training from scratch is not yet competitive, and MGM relies on a reasonable initialization.
Third, a principled adaptation of classifier-free guidance to one-step MGM, which is important for text-to-image generation, remains an open question.

% \vspace{-2mm}
% \section{Discussion and Limitations}\label{sec:discussion}
% \vspace{-2mm}

% We introduced \textit{Midpoint Generative Models} (MGM), a framework for training one-step generators by minimizing the generalized Midpoint Divergence. Our construction reframes symmetric stochastic interpolants, originally used to define generative paths, as a basis for comparing endpoint distributions. This yields a definite divergence and a practical variational objective for direct generator training.

% As MGM introduces a new training paradigm, several questions remain open. In particular, its performance may depend on the choice of stochastic interpolant, noise schedule, and interpolation-time sampling distribution during training. Understanding how these design choices affect optimization stability and sample quality is an important direction for future work.

% \textbf{Limitations.}
% Our practical training procedure uses a minimax objective with an adversarially trained displacement field. As in other adversarial training methods, this can introduce optimization instability and sensitivity to the balance between generator and displacement-field updates.
% \section*{Limitations}

\section*{Acknowledgments}
The work was supported by the grant for research centers in the field of AI provided by the Ministry of Economic Development of the Russian Federation in accordance with the agreement 000000C313925P4F0002 and the agreement No.~139-10-2025-033.

% \clearpage
\bibliographystyle{unsrtnat}
\bibliography{bibliography}

@inproceedings{manukhov2026interaction,
  title     = {Interaction Field Matching: Overcoming Limitations of Electrostatic Models},
  author    = {Manukhov, Stepan and Kolesov, Alexander and Palyulin, Vladimir V. and Korotin, Alexander},
  booktitle = {International Conference on Learning Representations},
  year      = {2026}
}

@inproceedings{xu2022poisson,
  title     = {Poisson Flow Generative Models},
  author    = {Xu, Yilun and Liu, Ziming and Tegmark, Max and Jaakkola, Tommi},
  booktitle = {Advances in Neural Information Processing Systems},
  year      = {2022}
}

@inproceedings{kolesov2025field,
  title     = {Field Matching: an Electrostatic Paradigm to Generate and Transfer Data},
  author    = {Kolesov, Alexander and Manukhov, Stepan and Palyulin, Vladimir V. and Korotin, Alexander},
  booktitle = {International Conference on Machine Learning},
  year      = {2025}
}

@inproceedings{shlenskii2026unlocking,
  title     = {Unlocking the Duality between Flow and Field Matching},
  author    = {Shlenskii, Daniil and Varlamov, Alexander and Buzun, Nazar and Korotin, Alexander},
  booktitle = {Advances in Neural Information Processing Systems},
  volume    = {39},
  year      = {2026}
}

@inproceedings{karras2024analyzing,
  title={Analyzing and improving the training dynamics of diffusion models},
  author={Karras, Tero and Aittala, Miika and Lehtinen, Jaakko and Hellsten, Janne and Aila, Timo and Laine, Samuli},
  booktitle={2024 IEEE/CVF Conference on Computer Vision and Pattern Recognition (CVPR)},
  pages={24174--24184},
  year={2024},
  organization={IEEE}
}

@inproceedings{sabour2026align,
  title={Align Your Flow: Scaling Continuous-Time Flow Map Distillation},
  author={Sabour, Amirmojtaba and Fidler, Sanja and Kreis, Karsten},
  booktitle={Advances in Neural Information Processing Systems},
  year={2025}
}

@article{wang2026zero,
  title={Zero-Flow Encoders},
  author={Wang, Yakun and Wang, Leyang and Liu, Song and Suzuki, Taiji},
  journal={arXiv preprint arXiv:2602.00797},
  year={2026}
}

@article{lipman2022flow,
  title={Flow matching for generative modeling},
  author={Lipman, Yaron and Chen, Ricky TQ and Ben-Hamu, Heli and Nickel, Maximilian and Le, Matt},
  journal={arXiv preprint arXiv:2210.02747},
  year={2022}
}

@article{liu2022flow,
  title={Flow straight and fast: Learning to generate and transfer data with rectified flow},
  author={Liu, Xingchao and Gong, Chengyue and Liu, Qiang},
  journal={arXiv preprint arXiv:2209.03003},
  year={2022}
}

@article{albergo2025stochastic,
  title={Stochastic interpolants: A unifying framework for flows and diffusions},
  author={Albergo, Michael and Boffi, Nicholas M and Vanden-Eijnden, Eric},
  journal={Journal of Machine Learning Research},
  volume={26},
  number={209},
  pages={1--80},
  year={2025}
}

@article{goodfellow2014generative,
  title={Generative adversarial nets},
  author={Goodfellow, Ian J and Pouget-Abadie, Jean and Mirza, Mehdi and Xu, Bing and Warde-Farley, David and Ozair, Sherjil and Courville, Aaron and Bengio, Yoshua},
  journal={Advances in neural information processing systems},
  volume={27},
  year={2014}
}

@article{yin2024improved,
  title={Improved distribution matching distillation for fast image synthesis},
  author={Yin, Tianwei and Gharbi, Micha{\"e}l and Park, Taesung and Zhang, Richard and Shechtman, Eli and Durand, Fredo and Freeman, Bill},
  journal={Advances in neural information processing systems},
  volume={37},
  pages={47455--47487},
  year={2024}
}

@inproceedings{zhou2024score,
  title={Score identity distillation: Exponentially fast distillation of pretrained diffusion models for one-step generation},
  author={Zhou, Mingyuan and Zheng, Huangjie and Wang, Zhendong and Yin, Mingzhang and Huang, Hai},
  booktitle={Forty-first International Conference on Machine Learning},
  year={2024}
}

@article{huang2024flow,
  title={Flow generator matching},
  author={Huang, Zemin and Geng, Zhengyang and Luo, Weijian and Qi, Guo-jun},
  journal={arXiv preprint arXiv:2410.19310},
  year={2024}
}

@inproceedings{liu2022let,
  title={Let us Build Bridges: Understanding and Extending Diffusion Generative Models},
  author={Liu, Xingchao and Wu, Lemeng and Ye, Mao and others},
    year={2022},
  booktitle={NeurIPS 2022 Workshop on Score-Based Methods}
}

@article{zheng2024diffusion,
  title={Diffusion bridge implicit models},
  author={Zheng, Kaiwen and He, Guande and Chen, Jianfei and Bao, Fan and Zhu, Jun},
  journal={arXiv preprint arXiv:2405.15885},
  year={2024}
}

@article{he2024consistency,
  title={Consistency diffusion bridge models},
  author={He, Guande and Zheng, Kaiwen and Chen, Jianfei and Bao, Fan and Zhu, Jun},
  journal={Advances in Neural Information Processing Systems},
  volume={37},
  pages={23516--23548},
  year={2024}
}

@article{wang2025implicit,
  title={Implicit Image-to-Image Schr{\"o}dinger Bridge for image restoration},
  author={Wang, Yuang and Yoon, Siyeop and Jin, Pengfei and Tivnan, Matthew and Song, Sifan and Chen, Zhennong and Hu, Rui and Zhang, Li and Li, Quanzheng and Chen, Zhiqiang and others},
  journal={Pattern Recognition},
  volume={165},
  pages={111627},
  year={2025},
  publisher={Elsevier}
}

@article{zhou2023denoising,
  title={Denoising diffusion bridge models},
  author={Zhou, Linqi and Lou, Aaron and Khanna, Samar and Ermon, Stefano},
  journal={arXiv preprint arXiv:2309.16948},
  year={2023}
}

@article{karras2022elucidating,
  title={Elucidating the design space of diffusion-based generative models},
  author={Karras, Tero and Aittala, Miika and Aila, Timo and Laine, Samuli},
  journal={Advances in neural information processing systems},
  volume={35},
  pages={26565--26577},
  year={2022}
}

@techreport{krizhevsky2009learning,
  title={Learning Multiple Layers of Features from Tiny Images},
  author={Krizhevsky, Alex},
  institution={University of Toronto},
  year={2009}
}

@inproceedings{deng2009imagenet,
  title={{ImageNet}: A Large-Scale Hierarchical Image Database},
  author={Deng, Jia and Dong, Wei and Socher, Richard and Li, Li-Jia and Li, Kai and Fei-Fei, Li},
  booktitle={IEEE Conference on Computer Vision and Pattern Recognition},
  year={2009}
}

@article{heusel2017gans,
  title={Gans trained by a two time-scale update rule converge to a local nash equilibrium},
  author={Heusel, Martin and Ramsauer, Hubert and Unterthiner, Thomas and Nessler, Bernhard and Hochreiter, Sepp},
  journal={Advances in neural information processing systems},
  volume={30},
  year={2017}
}

@inproceedings{
albergo2023building,
title={Building Normalizing Flows with Stochastic Interpolants},
author={Michael Samuel Albergo and Eric Vanden-Eijnden},
booktitle={The Eleventh International Conference on Learning Representations },
year={2023},
url={https://openreview.net/forum?id=li7qeBbCR1t}
}

@article{song2020denoising,
  title={Denoising diffusion implicit models},
  author={Song, Jiaming and Meng, Chenlin and Ermon, Stefano},
  journal={arXiv preprint arXiv:2010.02502},
  year={2020}
}

@article{lu2022dpm,
  title={Dpm-solver: A fast ode solver for diffusion probabilistic model sampling in around 10 steps},
  author={Lu, Cheng and Zhou, Yuhao and Bao, Fan and Chen, Jianfei and Li, Chongxuan and Zhu, Jun},
  journal={Advances in neural information processing systems},
  volume={35},
  pages={5775--5787},
  year={2022}
}

@article{geng2025mean,
  title={Mean flows for one-step generative modeling},
  author={Geng, Zhengyang and Deng, Mingyang and Bai, Xingjian and Kolter, J Zico and He, Kaiming},
  journal={arXiv preprint arXiv:2505.13447},
  year={2025}
}

@article{zhou2025inductive,
  title={Inductive moment matching},
  author={Zhou, Linqi and Ermon, Stefano and Song, Jiaming},
  journal={arXiv preprint arXiv:2503.07565},
  year={2025}
}

@article{dhariwal2021diffusion,
  title={Diffusion models beat gans on image synthesis},
  author={Dhariwal, Prafulla and Nichol, Alexander},
  journal={Advances in neural information processing systems},
  volume={34},
  pages={8780--8794},
  year={2021}
}

@inproceedings{rombach2022high,
  title={High-resolution image synthesis with latent diffusion models},
  author={Rombach, Robin and Blattmann, Andreas and Lorenz, Dominik and Esser, Patrick and Ommer, Bj{\"o}rn},
  booktitle={Proceedings of the IEEE/CVF conference on computer vision and pattern recognition},
  pages={10684--10695},
  year={2022}
}

@article{ho2022video,
  title={Video diffusion models},
  author={Ho, Jonathan and Salimans, Tim and Gritsenko, Alexey and Chan, William and Norouzi, Mohammad and Fleet, David J},
  journal={Advances in neural information processing systems},
  volume={35},
  pages={8633--8646},
  year={2022}
}

@inproceedings{
kong2021diffwave,
title={DiffWave: A Versatile Diffusion Model for Audio Synthesis},
author={Zhifeng Kong and Wei Ping and Jiaji Huang and Kexin Zhao and Bryan Catanzaro},
booktitle={International Conference on Learning Representations},
year={2021},
url={https://openreview.net/forum?id=a-xFK8Ymz5J}
}

@inproceedings{
meng2022sdedit,
title={{SDE}dit: Guided Image Synthesis and Editing with Stochastic Differential Equations},
author={Chenlin Meng and Yutong He and Yang Song and Jiaming Song and Jiajun Wu and Jun-Yan Zhu and Stefano Ermon},
booktitle={International Conference on Learning Representations},
year={2022},
url={https://openreview.net/forum?id=aBsCjcPu_tE}
}

@inproceedings{saharia2022palette,
  title={Palette: Image-to-image diffusion models},
  author={Saharia, Chitwan and Chan, William and Chang, Huiwen and Lee, Chris and Ho, Jonathan and Salimans, Tim and Fleet, David and Norouzi, Mohammad},
  booktitle={ACM SIGGRAPH 2022 conference proceedings},
  pages={1--10},
  year={2022}
}

@inproceedings{
salimans2022progressive,
title={Progressive Distillation for Fast Sampling of Diffusion Models},
author={Tim Salimans and Jonathan Ho},
booktitle={International Conference on Learning Representations},
year={2022},
url={https://openreview.net/forum?id=TIdIXIpzhoI}
}

@inproceedings{song2023consistency,
  title={Consistency Models},
  author={Song, Yang and Dhariwal, Prafulla and Chen, Mark and Sutskever, Ilya},
  booktitle={International Conference on Machine Learning},
  pages={32211--32252},
  year={2023},
  organization={PMLR}
}

@inproceedings{yin2024one,
  title={One-step diffusion with distribution matching distillation},
  author={Yin, Tianwei and Gharbi, Micha{\"e}l and Zhang, Richard and Shechtman, Eli and Durand, Fredo and Freeman, William T and Park, Taesung},
  booktitle={Proceedings of the IEEE/CVF conference on computer vision and pattern recognition},
  pages={6613--6623},
  year={2024}
}

@inproceedings{sauer2024adversarial,
  title={Adversarial diffusion distillation},
  author={Sauer, Axel and Lorenz, Dominik and Blattmann, Andreas and Rombach, Robin},
  booktitle={European Conference on Computer Vision},
  pages={87--103},
  year={2024},
  organization={Springer}
}

@inproceedings{
gu2023boot,
title={{BOOT}: Data-free Distillation of Denoising Diffusion Models with Bootstrapping},
author={Jiatao Gu and Shuangfei Zhai and Yizhe Zhang and Lingjie Liu and Joshua M. Susskind},
booktitle={ICML 2023 Workshop on Structured Probabilistic Inference {\&} Generative Modeling},
year={2023},
url={https://openreview.net/forum?id=ZeM7S01Xi8}
}

@inproceedings{
song2024improved,
title={Improved Techniques for Training Consistency Models},
author={Yang Song and Prafulla Dhariwal},
booktitle={The Twelfth International Conference on Learning Representations},
year={2024},
url={https://openreview.net/forum?id=WNzy9bRDvG}
}

@inproceedings{
lu2025simplifying,
title={Simplifying, Stabilizing and Scaling Continuous-time Consistency Models},
author={Cheng Lu and Yang Song},
booktitle={The Thirteenth International Conference on Learning Representations},
year={2025},
url={https://openreview.net/forum?id=LyJi5ugyJx}
}

@inproceedings{
frans2025one,
title={One Step Diffusion via Shortcut Models},
author={Kevin Frans and Danijar Hafner and Sergey Levine and Pieter Abbeel},
booktitle={The Thirteenth International Conference on Learning Representations},
year={2025},
url={https://openreview.net/forum?id=OlzB6LnXcS}
}

@article{luo2023diffinstruct,
  title={Diff-instruct: A universal approach for transferring knowledge from pre-trained diffusion models},
  author={Luo, Weijian and Hu, Tianyang and Zhang, Shifeng and Sun, Jiacheng and Li, Zhenguo and Zhang, Zhihua},
  journal={Advances in Neural Information Processing Systems},
  volume={36},
  pages={76525--76546},
  year={2023}
}

@article{xu2025fdistill,
  title={One-step Diffusion Models with $ f $-Divergence Distribution Matching},
  author={Xu, Yilun and Nie, Weili and Vahdat, Arash},
  journal={arXiv preprint arXiv:2502.15681},
  year={2025}
}

@article{luo2024sim,
  title={One-step diffusion distillation through score implicit matching},
  author={Luo, Weijian and Huang, Zemin and Geng, Zhengyang and Kolter, J Zico and Qi, Guo-jun},
  journal={Advances in Neural Information Processing Systems},
  volume={37},
  pages={115377--115408},
  year={2024}
}

@inproceedings{
kornilov2025universal,
title={Universal Inverse Distillation for Matching Models with Real-Data Supervision (No {GAN}s)},
author={Nikita Maksimovich Kornilov and David Li and Tikhon Mavrin and Aleksei Leonov and Nikita Gushchin and Evgeny Burnaev and Iaroslav Sergeevich Koshelev and Alexander Korotin},
booktitle={The Fourteenth International Conference on Learning Representations},
year={2026},
url={https://openreview.net/forum?id=8NuN5UzXLC}
}

@article{
boffi2025flowmap,
title={Flow map matching with stochastic interpolants: A mathematical framework for consistency models},
author={Nicholas Matthew Boffi and Michael Samuel Albergo and Eric Vanden-Eijnden},
journal={Transactions on Machine Learning Research},
issn={2835-8856},
year={2025},
url={https://openreview.net/forum?id=cqDH0e6ak2},
note={}
}

@article{hu2026cmt,
  title={Cmt: Mid-training for efficient learning of consistency, mean flow, and flow map models},
  author={Hu, Zheyuan and Lai, Chieh-Hsin and Mitsufuji, Yuki and Ermon, Stefano},
  journal={arXiv preprint arXiv:2509.24526},
  year={2025}
}

@article{nowozin2016fgan,
  title={f-gan: Training generative neural samplers using variational divergence minimization},
  author={Nowozin, Sebastian and Cseke, Botond and Tomioka, Ryota},
  journal={Advances in neural information processing systems},
  volume={29},
  year={2016}
}

@inproceedings{arjovsky2017wasserstein,
  title={Wasserstein generative adversarial networks},
  author={Arjovsky, Martin and Chintala, Soumith and Bottou, L{\'e}on},
  booktitle={International conference on machine learning},
  pages={214--223},
  year={2017},
  organization={Pmlr}
}

@inproceedings{
xiao2021denoising,
title={Tackling the Generative Learning Trilemma with Denoising Diffusion {GAN}s},
author={Zhisheng Xiao and Karsten Kreis and Arash Vahdat},
booktitle={International Conference on Learning Representations},
year={2022},
url={https://openreview.net/forum?id=JprM0p-q0Co}
}

@inproceedings{
kim2024consistency,
title={Consistency Trajectory Models: Learning Probability Flow {ODE} Trajectory of Diffusion},
author={Dongjun Kim and Chieh-Hsin Lai and Wei-Hsiang Liao and Naoki Murata and Yuhta Takida and Toshimitsu Uesaka and Yutong He and Yuki Mitsufuji and Stefano Ermon},
booktitle={The Twelfth International Conference on Learning Representations},
year={2024},
url={https://openreview.net/forum?id=ymjI8feDTD}
}

@inproceedings{
wang2023diffusiongan,
title={Diffusion-{GAN}: Training {GAN}s with Diffusion},
author={Zhendong Wang and Huangjie Zheng and Pengcheng He and Weizhu Chen and Mingyuan Zhou},
booktitle={The Eleventh International Conference on Learning Representations },
year={2023},
url={https://openreview.net/forum?id=HZf7UbpWHuA}
}

@article{berthelot2023tract,
  title={Tract: Denoising diffusion models with transitive closure time-distillation},
  author={Berthelot, David and Autef, Arnaud and Lin, Jierui and Yap, Dian Ang and Zhai, Shuangfei and Hu, Siyuan and Zheng, Daniel and Talbott, Walter and Gu, Eric},
  journal={arXiv preprint arXiv:2303.04248},
  year={2023}
}

@inproceedings{zheng2023fast,
  title={Fast sampling of diffusion models via operator learning},
  author={Zheng, Hongkai and Nie, Weili and Vahdat, Arash and Azizzadenesheli, Kamyar and Anandkumar, Anima},
  booktitle={International conference on machine learning},
  pages={42390--42402},
  year={2023},
  organization={PMLR}
}

@inproceedings{
geng2025consistency,
title={Consistency Models Made Easy},
author={Zhengyang Geng and Ashwini Pokle and Weijian Luo and Justin Lin and J Zico Kolter},
booktitle={The Thirteenth International Conference on Learning Representations},
year={2025},
url={https://openreview.net/forum?id=xQVxo9dSID}
}

@inproceedings{
wang2025stable,
title={Stable Consistency Tuning: Understanding and Improving Consistency Models},
author={Fu-Yun Wang and Zhengyang Geng and Hongsheng Li},
booktitle={ICLR 2025 Workshop on Deep Generative Model in Machine Learning: Theory, Principle and Efficacy},
year={2025},
url={https://openreview.net/forum?id=5RoPe2ShXx}
}

@inproceedings{
silvestri2025vct,
title={{VCT}: Training Consistency Models with Variational Noise Coupling},
author={Gianluigi Silvestri and Luca Ambrogioni and Chieh-Hsin Lai and Yuhta Takida and Yuki Mitsufuji},
booktitle={Forty-second International Conference on Machine Learning},
year={2025},
url={https://openreview.net/forum?id=CMoX0BEsDs}
}

@inproceedings{lee2025truncated,
title={Truncated Consistency Models},
author={Sangyun Lee and Yilun Xu and Tomas Geffner and Giulia Fanti and Karsten Kreis and Arash Vahdat and Weili Nie},
booktitle={The Thirteenth International Conference on Learning Representations},
year={2025},
url={https://openreview.net/forum?id=ZYDEJEvCbv}
}

@inproceedings{
gushchin2025inverse,
title={Inverse Bridge Matching Distillation},
author={Nikita Gushchin and David Li and Daniil Selikhanovych and Evgeny Burnaev and Dmitry Baranchuk and Alexander Korotin},
booktitle={Forty-second International Conference on Machine Learning},
year={2025},
url={https://openreview.net/forum?id=UCJSF6Vt0C}
}

@article{shlenskii2025overclocking,
  title={Overclocking Electrostatic Generative Models},
  author={Shlenskii, Daniil and Korotin, Alexander},
  booktitle={Forty-third International Conference on Machine Learning},
  year={2026},
  url={https://openreview.net/forum?id=flo449mncA}
}

@article{karras2020training,
  title={Training generative adversarial networks with limited data},
  author={Karras, Tero and Aittala, Miika and Hellsten, Janne and Laine, Samuli and Lehtinen, Jaakko and Aila, Timo},
  journal={Advances in neural information processing systems},
  volume={33},
  pages={12104--12114},
  year={2020}
}

@article{chen2025diffratio,
  title={DiffRatio: Training One-Step Diffusion Models Without Teacher Supervision},
  author={Chen, Wenlin and Zhang, Mingtian and He, Jiajun and Ou, Zijing and Hern{\'a}ndez-Lobato, Jos{\'e} Miguel and Sch{\"o}lkopf, Bernhard and Barber, David},
  journal={arXiv preprint arXiv:2502.08005},
  year={2025}
}

%%%%%%%%%%%%%%%%%%%%%%%%%%%%%%%%%%%%%%%%%%%%%%%%%%%%%%%%%%%%

\newpage
\appendix
\section{Proofs}\label{app:appendix}
In this section, we provide proofs for all theorems and propositions in the main part.
\subsection{Proof of Proposition~\ref{prop:midpoint_symmetry}}

Let
\[
    \Delta := X_1 - X_0,
    \qquad
    M := X_{1/2} = \frac{X_0+X_1}{2}.
\]
Since \(X_0\) and \(X_1\) are independent samples from the same distribution,
\[
    (X_0,X_1) \stackrel{d}{=} (X_1,X_0).
\]
Under this exchange, \(M\) is unchanged while \(\Delta\) changes sign. Hence
\[
    (\Delta,M) \stackrel{d}{=} (-\Delta,M).
\]
Therefore the conditional law of \(\Delta\) given \(M\) is symmetric about the origin, and so
\[
    \mathbb{E}[\Delta \mid M] = 0 \quad \text{a.s.}
\]
Since $\Delta = X_1 - X_0$ and \(M=X_{1/2}\), this is exactly
\[
    v_{1/2}(x)
    =
    \mathbb{E}[X_1-X_0 \mid X_{1/2}=x]
    =
    0
\]
for \(p_{1/2}\)-almost every \(x\).

\subsection{Proof of Theorem~\ref{thm:definiteness_of_midpoint_divergence}}
Let
\[
    \Delta := X_1-X_0,
    \qquad
    M := X_{1/2}=\frac{X_0+X_1}{2}.
\]
Then, by \new{Definition~\ref{def:midpoint_divegence}},
\[
    D_{\mathrm{mid}}(p_0,p_1)
    =
    \mathbb{E}\bigl[
        \|\mathbb{E}[\Delta \mid M]\|_2^2
    \bigr].
\]

If \(p_0=p_1\), then Proposition~\ref{prop:midpoint_symmetry} gives
\[
    \mathbb{E}[\Delta \mid M]=0
    \quad \text{a.s.}
\]
and hence \(D_{\mathrm{mid}}(p_0,p_1)=0\).

Conversely, suppose that
\[
    D_{\mathrm{mid}}(p_0,p_1)=0.
\]
Since the integrand is nonnegative,
\[
    \mathbb{E}[\Delta \mid M]=0
    \quad \text{a.s.}
\]
\new{
    Since $X_0$ and $X_1$ are assumed to have sub-exponential tails,
    i.e., \(\mathbb{E}\,e^{\alpha\|X_j\|_2}<\infty\) for \(j\in\{0,1\}\) and some \(\alpha>0\),
    \(X_0\) and \(X_1\) have finite moments of all orders, so \(\Delta\) is integrable.
    Thus, by the tower property, for every measurable function \(\varphi\) such that \(\Delta\,\varphi(M)\) is integrable, we have
}
\[
    \mathbb{E}[\Delta \, \varphi(M)] = 0.
\]
\new{For every \(u\in\mathbb{R}^d\) with \(\|u\|_2<2\alpha\) (we take $\alpha$ that works for both $X_0$ and $X_1$),} the function
\[
    \varphi_u(m) := e^{\langle u,m\rangle}
\]
\new{satisfies \(\|\Delta\,\varphi_u(M)\|_2\le(\|X_0\|_2+\|X_1\|_2)\,e^{\|u\|_2\|X_0\|_2/2}\,e^{\|u\|_2\|X_1\|_2/2}\). Since \(\|x\|_2\,e^{\|u\|_2\|x\|_2/2}\le C\,e^{\alpha\|x\|_2}\) for some constant \(C\), the right-hand side is integrable by independence of \(X_0\) and \(X_1\).}
\new{Therefore, applying the identity above with \(\varphi=\varphi_u\) and recalling that \(M=(X_0+X_1)/2\),}
\[
    \mathbb{E}
    \left[
        (X_1-X_0)
        e^{\langle u,(X_0+X_1)/2\rangle}
    \right]
    =
    0.
\]
Using independence of \(X_0\) and \(X_1\), this becomes
\[
    \mathbb{E}\!\left[X_1 e^{\langle u,X_1\rangle/2}\right]
    \mathbb{E}\!\left[e^{\langle u,X_0\rangle/2}\right]
    =
    \mathbb{E}\!\left[X_0 e^{\langle u,X_0\rangle/2}\right]
    \mathbb{E}\!\left[e^{\langle u,X_1\rangle/2}\right].
\]
Define the moment generating functions
\[
    G_j(v) := \mathbb{E}\!\left[e^{\langle v,X_j\rangle}\right],
    \qquad j\in\{0,1\}.
\]
\new{For \(\|v\|_2<\alpha\), we have \(e^{\langle v,x\rangle}\le e^{\alpha\|x\|_2}\), so \(G_0\) and \(G_1\) are finite on the ball \(B_\alpha:=\{v\in\mathbb{R}^d:\|v\|_2<\alpha\}\). Moreover, for every \(a<\alpha\), the gradient \(x\,e^{\langle v,x\rangle}\) is dominated, uniformly over \(\|v\|_2\le a\), by the integrable function \(C_a\,e^{\alpha\|x\|_2}\); hence differentiation under expectation is justified, and on \(B_\alpha\)}
\[
    \nabla G_j(v)
    =
    \mathbb{E}\!\left[X_j e^{\langle v,X_j\rangle}\right].
\]
Setting \(v=u/2\), the preceding identity yields
\[
    G_0(v)\nabla G_1(v)
    =
    G_1(v)\nabla G_0(v)
    \qquad
    \text{for all \new{$v\in B_\alpha$}}.
\]
Since \(G_0(v)>0\) and \(G_1(v)>0\),
\[
    \nabla \log G_1(v)
    =
    \nabla \log G_0(v).
\]
Hence \(\log G_1-\log G_0\) is constant on \new{the connected set \(B_\alpha\)}. Evaluating at \(v=0\) gives
\[
    G_0(0)=G_1(0)=1,
\]
so the constant is zero. Thus
\[
    G_0(v)=G_1(v)
    \qquad
    \text{for all \new{$v\in B_\alpha$}}.
\]
By \new{the uniqueness theorem for moment generating functions finite on a neighborhood of the origin, we obtain} \(p_0=p_1\). Therefore
\[
    D_{\mathrm{mid}}(p_0,p_1)=0
    \quad\Longleftrightarrow\quad
    p_0=p_1.
\]

\subsection{Proof of Proposition~\ref{prop:flip_symmetry}}
Fix \(t\in[0,1]\) and let
\[
    \Delta := X_1-X_0.
\]
We need to show that
\[
    \mathbb{E}[\Delta \mid \widetilde X_t]=0.
\]

Since \(X_0\) and \(X_1\) are independent samples from the same distribution and
\(B\sim\mathrm{Bernoulli}(1/2)\) is independent of them,
\[
    (X_0,X_1,B)
    \stackrel{d}{=}
    (X_1,X_0,1-B).
\]
Under this transformation, \(\Delta\) changes sign. On the other hand, the flipped observation
\[
    \widetilde X_t
    =
    (1-B)\bigl((1-t)X_0+tX_1\bigr)
    +
    B\bigl(tX_0+(1-t)X_1\bigr)
\]
is unchanged. Therefore
\[
    (\Delta,\widetilde X_t)
    \stackrel{d}{=}
    (-\Delta,\widetilde X_t).
\]
It follows that the conditional law of \(\Delta\) given \(\widetilde X_t\) is symmetric about the origin, and hence
\[
    \mathbb{E}[\Delta \mid \widetilde X_t]=0
    \quad \text{a.s.}
\]
Equivalently,
\[
    \widetilde v_t(x)
    =
    \mathbb{E}[X_1-X_0 \mid \widetilde X_t=x]
    =
    0
\]
for \(\widetilde p_t\)-almost every \(x\).

\subsection{Proof of Theorem~\ref{thm:time_intergrated_midpoint_divergence}}
Theorem~\ref{thm:time_intergrated_midpoint_divergence} is the deterministic linear interpolant special case of Theorem~\ref{thm:generalized_midpoint_divergence}.
Indeed, take
\[
    I_t(x_0,x_1) := (1-t)x_0 + t x_1,
    \qquad
    \sigma_t := 0.
\]
Then the flipped stochastic interpolant
in~\eqref{eq:flipped_stochastic_interpolant}
reduces to the flipped linear interpolation in~\eqref{eq:flip_flow_matching_interpolant}:
\[
    \widetilde X_t
    =
    \begin{cases}
        (1-t)X_0+tX_1, & B=0,\\
        tX_0+(1-t)X_1, & B=1.
    \end{cases}
\]
Consequently,
\[
    D_{t\text{-}\mathrm{mid}}^{I,\sigma}(p_0,p_1)
    =
    D_{t\text{-}\mathrm{mid}}(p_0,p_1).
\]
The claim therefore follows directly from Theorem~\ref{thm:generalized_midpoint_divergence}.

\subsection{Proof of Theorem~\ref{thm:generalized_midpoint_divergence}}
\label{app:generalized_midpoint_divergence_proof}
Let
\[
    \Delta := X_1-X_0,
    \qquad
    m_t(x) := \mathbb{E}[\Delta \mid \widetilde X_t=x].
\]
By definition,
\[
    D_{t\text{-}\mathrm{mid}}^{I,\sigma}(p_0,p_1)
    =
    \int_0^{1/2}
    \mathbb{E}\bigl[\|m_t(\widetilde X_t)\|_2^2\bigr]\,dt.
\]
Nonnegativity is immediate.

First suppose that \(p_0=p_1\). Then \(X_0\) and \(X_1\) are exchangeable. Since
\(B\sim\mathrm{Bernoulli}(1/2)\) is independent of \((X_0,X_1,\epsilon)\), the transformation
\[
    (X_0,X_1,B,\epsilon)
    \mapsto
    (X_1,X_0,1-B,\epsilon)
\]
preserves the joint distribution. Under this transformation, \(\Delta\) changes sign. On the other
hand, the flipped observation~\eqref{eq:flipped_stochastic_interpolant}, which due to symmetry could be rewritten as
\[
    \widetilde X_t
    =
    \begin{cases}
        I_t(X_0,X_1)+\sigma_t\epsilon, & B=0,\\
        I_t(X_1,X_0)+\sigma_t\epsilon, & B=1,
    \end{cases}
\]
is unchanged. Hence
\[
    (\Delta,\widetilde X_t)
    \stackrel{d}{=}
    (-\Delta,\widetilde X_t).
\]
Therefore the conditional law of \(\Delta\) given \(\widetilde X_t\) is symmetric about the origin, so
\[
    m_t(\widetilde X_t)
    =
    \mathbb{E}[\Delta \mid \widetilde X_t]
    =
    0
    \quad \text{a.s.}
\]
for every \(t\in[0,1/2]\). Thus
\[
    D_{t\text{-}\mathrm{mid}}^{I,\sigma}(p_0,p_1)=0.
\]

Conversely, suppose that
\[
    D_{t\text{-}\mathrm{mid}}^{I,\sigma}(p_0,p_1)=0.
\]
Since the integrand is nonnegative,
\[
    m_t(\widetilde X_t)
    =
    \mathbb{E}[\Delta \mid \widetilde X_t]
    =
    0
    \quad \text{a.s.}
\]
for almost every \(t\in[0,1/2]\). Choose a sequence \(t_n\downarrow 0\) from this full-measure set.
By the endpoint continuity of the stochastic interpolant,
\[
    \widetilde X_{t_n}
    \longrightarrow
    \new{Y}
    \quad \text{a.s.},
\]
where
\[
    \new{Y} :=
    \begin{cases}
        X_0, & B=0,\\
        X_1, & B=1.
    \end{cases}
\]
Let \(\varphi:\mathbb{R}^d\to\mathbb{R}\) be bounded and continuous.
For each \(n\), since
\[
    m_{t_n}(\widetilde X_{t_n})
    =
    \mathbb{E}[\Delta \mid \widetilde X_{t_n}]
    =
    0
    \quad \text{a.s.},
\]
we have, by the tower property,
\[
\begin{aligned}
    \mathbb{E}\bigl[\Delta\,\varphi(\widetilde X_{t_n})\bigr]
    &=
    \mathbb{E}\bigl[
        \mathbb{E}[\Delta \mid \widetilde X_{t_n}]
        \varphi(\widetilde X_{t_n})
    \bigr] \\
    &=
    \mathbb{E}\bigl[
        m_{t_n}(\widetilde X_{t_n})
        \varphi(\widetilde X_{t_n})
    \bigr] \\
    &=
    0.
\end{aligned}
\]
Since \(\Delta\in L^1\) and \(\varphi\) is bounded, dominated convergence gives
\[
    \mathbb{E}[\Delta\,\varphi(\new{Y})] = 0.
\]
By a monotone-class argument, the same identity holds for every bounded Borel function
\(\varphi\).

Expanding over the two values of \(B\), we obtain
\[
    0
    =
    \frac12
    \mathbb{E}\bigl[(X_1-X_0)\varphi(X_0)\bigr]
    +
    \frac12
    \mathbb{E}\bigl[(X_1-X_0)\varphi(X_1)\bigr].
\]
Let
\[
    \bar x_0 := \mathbb{E}[X_0],
    \qquad
    \bar x_1 := \mathbb{E}[X_1].
\]
Using independence of \(X_0\) and \(X_1\), the previous identity becomes
\[
    \int \varphi(x)(\bar x_1-x)\,p_0(dx)
    +
    \int \varphi(x)(x-\bar x_0)\,p_1(dx)
    =
    0
\]
for every bounded Borel \(\varphi\). Taking \(\varphi\equiv 1\) gives
\[
    \bar x_0=\bar x_1.
\]
Denote the common value by \(\bar x\). Then
\[
    \int \varphi(x)(x-\bar x)\,(p_1-p_0)(dx)
    =
    0
\]
for every bounded Borel \(\varphi\).

Let
\[
    \mu := p_1-p_0.
\]
We have shown that
\[
    (x-\bar x)\,\mu(dx)=0
\]
as a vector-valued signed measure. We now show that this implies \(\mu=0\).

Fix a coordinate \(j\in\{1,\dots,d\}\) and an integer \(n\ge 1\), and define
\[
    A_{j,n}
    :=
    \{x\in\mathbb{R}^d:\ |x_j-\bar x_j|\ge 1/n\}.
\]
For any Borel set \(A\subseteq A_{j,n}\), the function
\[
    \varphi(x)
    :=
    \frac{\mathbf{1}_A(x)}{x_j-\bar x_j}
\]
is bounded and Borel. Applying the \(j\)-th coordinate of the signed-measure identity gives
\[
    \mu(A)
    =
    \int \varphi(x)(x_j-\bar x_j)\,\mu(dx)
    =
    0.
\]
Thus \(\mu\) vanishes on every Borel subset of \(A_{j,n}\). Since
\[
    \mathbb{R}^d\setminus\{\bar x\}
    =
    \bigcup_{j=1}^d \bigcup_{n=1}^\infty A_{j,n},
\]
we have
\[
    \mu(\mathbb{R}^d\setminus\{\bar x\})=0.
\]
Hence \(\mu\) is supported on \(\{\bar x\}\). But \(\mu\) is the difference of two probability measures,
so
\[
    \mu(\mathbb{R}^d)=p_1(\mathbb{R}^d)-p_0(\mathbb{R}^d)=0.
\]
Therefore \(\mu(\{\bar x\})=0\), and consequently \(\mu\equiv 0\). Hence \(p_1\equiv p_0\).

Combining both directions,
\[
    D_{t\text{-}\mathrm{mid}}^{I,\sigma}(p_0,p_1)=0
    \quad\Longleftrightarrow\quad
    p_0=p_1.
\]

\subsection{Proof of Proposition~\ref{prop:variation_midpoint_objective}}
Let
\[
    \Delta := X_1-X_0,
    \qquad
    m_t(x) := \mathbb{E}[\Delta \mid \widetilde X_t=x].
\]
Then
\[
    D_{t\text{-}\mathrm{mid}}^{I,\sigma}(p_\theta,p_{\mathrm{data}})
    =
    \int_0^{1/2}
    \mathbb{E}\bigl[\|m_t(\widetilde X_t)\|_2^2\bigr]\,dt.
\]

Fix \(t\in[0,1/2]\) and let \(f_t\) be any square-integrable vector-valued function. Recall that
\[
    m_t(\widetilde X_t)
    =
    \mathbb{E}[\Delta \mid \widetilde X_t].
\]
Since \(f_t(\widetilde X_t)\) is measurable with respect to \(\widetilde X_t\), the tower property gives
\[
\begin{aligned}
    \mathbb{E}\bigl[
        \langle f_t(\widetilde X_t),\Delta\rangle
    \bigr]
    &=
    \mathbb{E}\Bigl[
        \mathbb{E}\bigl[
            \langle f_t(\widetilde X_t),\Delta\rangle
            \mid \widetilde X_t
        \bigr]
    \Bigr] \\
    &=
    \mathbb{E}\Bigl[
        \left\langle
            f_t(\widetilde X_t),
            \mathbb{E}[\Delta \mid \widetilde X_t]
        \right\rangle
    \Bigr] \\
    &=
    \mathbb{E}\bigl[
        \langle f_t(\widetilde X_t),m_t(\widetilde X_t)\rangle
    \bigr].
\end{aligned}
\]
The second term does not involve \(\Delta\), so it is unchanged.
Therefore
\[
\begin{aligned}
    \mathbb{E}
    \left[
        2\langle f_t(\widetilde X_t),\Delta\rangle
        -
        \|f_t(\widetilde X_t)\|_2^2
    \right]
    =
    \mathbb{E}
    \left[
        2\langle f_t(\widetilde X_t),m_t(\widetilde X_t)\rangle
        -
        \|f_t(\widetilde X_t)\|_2^2
    \right].
\end{aligned}
\]
Completing the square,
\[
    2\langle f_t,m_t\rangle-\|f_t\|_2^2
    =
    \|m_t\|_2^2-\|f_t-m_t\|_2^2.
\]
Therefore
\[
\begin{aligned}
    \mathbb{E}
    \left[
        2\langle f_t(\widetilde X_t),\Delta\rangle
        -
        \|f_t(\widetilde X_t)\|_2^2
    \right]
    =
    \mathbb{E}\bigl[\|m_t(\widetilde X_t)\|_2^2\bigr]
    -
    \mathbb{E}\bigl[
        \|f_t(\widetilde X_t)-m_t(\widetilde X_t)\|_2^2
    \bigr].
\end{aligned}
\]
The first term is nonnegative and does not depend on $f_t$, so the expression is maximized when
\[
    f_t(\widetilde X_t)=m_t(\widetilde X_t)
    \quad \text{a.s.}
\]
Equivalently,
\[
    f_t^\star(x)
    =
    m_t(x)
    =
    \mathbb{E}[X_1-X_0 \mid \widetilde X_t=x]
\]
for \(\widetilde p_t\)-almost every \(x\).

Integrating the pointwise variational identity over \(t\in[0,1/2]\) yields
\[
\begin{aligned}
    D_{t\text{-}\mathrm{mid}}^{I,\sigma}(p_\theta,p_{\mathrm{data}})
    &=
    \max_{f_t}
    \int_0^{1/2}
    \mathbb{E}
    \left[
        2\langle f_t(\widetilde X_t),X_1-X_0\rangle
        -
        \|f_t(\widetilde X_t)\|_2^2
    \right]dt.
\end{aligned}
\]
This proves the variational representation and identifies the maximizer.

\section{Algorithm}\label{app:algorithm}

In this section, we provide a detailed description of the Midpoint Generative Models training procedure.
Algorithm~\ref{alg:mgm_training_algo} gives pseudocode, while
Algorithm~\ref{alg:mgm_algo_code} provides a corresponding Python implementation sketch.

\new{\textbf{Critic-only noise.} In this setting, the generator loss uses the randomly flipped interpolant with its noise term removed, while the \critic\ loss retains the chosen noise schedule. In Algorithm~\ref{alg:mgm_algo_code}, this corresponds to passing \texttt{sigma=0} to \texttt{mgm\_loss\_G} and the chosen \texttt{sigma} to \texttt{mgm\_loss\_f}.}

\textbf{Simpler variants of our objective.} The same training template can also be adapted to simpler variants of our objective,
including the midpoint-only divergence
$D_{\mathrm{mid}}$ in~\eqref{eq:midpoint_divegence}
and the naive unflipped time-integrated objective
$D_{\mathrm{t\text{-}mid}}^{\mathrm{naive}}$ in~\eqref{eq:naive_time_integrated_midpoint_divergence}.
\new{To recover these functionals exactly, we use the linear interpolant $I_t(x_0,x_1)=(1-t)x_0+tx_1$ and set $\sigma_t\equiv0$ in both losses.}
To obtain the midpoint-only variant, one fixes $t=1/2$ throughout training instead of sampling
$t\sim\mathcal U[0,1/2]$.
To obtain the naive unflipped time-integrated variant, one removes the random flip by setting
$B\equiv 0$ and samples $t$ over the full interval $[0,1]$ rather than $[0,1/2]$.
\new{Retaining nonzero noise yields the corresponding stochastic training variants.}
\begin{algorithm}[t]
\caption{MGM training procedure}
\label{alg:mgm_training_algo}
\begin{algorithmic}[1]
\Require
Generator $G_\theta$,
\critic\ $f_\psi$,
symmetric interpolant $(I_t,\sigma_t)$
\Repeat
    \State \textit{\# Update \critic\ $f_\psi$}
    \State Sample
    $z\sim p_z$,
    $x_1\sim p_{\rm data}$,
    $t\sim\mathcal{U}[0,1/2]$,
    $b\sim{\rm Bernoulli}(1/2)$,
    $\epsilon\sim\mathcal{N}(0,\mathrm{Id})$.
    \State Set $x_0 = G_\theta(z)$.
    \State Define
    \[
        x_t = I_t(x_0, x_1) + \sigma_t \epsilon,
        \qquad
        x_{1-t} = I_{1-t}(x_0, x_1) + \sigma_{1-t} \epsilon.
    \]
    \State Set the flipped observation
    $\tilde x_t = (1-b)x_t + b x_{1-t}$.
    \State Update $\psi$ by a gradient \new{descent} step on
    \[
        \hat{L}_\psi = \|f_\psi(\tilde x_t, t) - (x_1 - x_0)\|_2^2.
    \]
    \State \textit{\# Update generator $G_\theta$}
    \State Sample a fresh batch and recompute
    $x_0$, $x_1$, and $\tilde{x}_t$ as above.
    \State Update $\theta$ by a gradient \new{descent} step on
    \[
        \hat{L}_\theta =
        2 \langle
            f_\psi(\tilde x_t, t), x_1 - x_0
        \rangle
        -
        \|f_\psi(\tilde x_t, t)\|_2^2.
    \]
\Until{convergence}
\State \textbf{Return} $G_\theta$
\end{algorithmic}
\end{algorithm}

\definecolor{keywordblue}{RGB}{0,0,245}      % def, return, ...
\definecolor{commentteal}{RGB}{64,128,128}   % # comments
\definecolor{funcpurple}{RGB}{205,85,142}    % helper functions

\lstset{
  language=Python,
  basicstyle=\ttfamily\small,
  keywordstyle=\bfseries\color{keywordblue},
  commentstyle=\color{commentteal},
  emphstyle=\color{funcpurple},
  emph={sample_noise, sample_data, rand, randn, stop_grad, sqrt,
        float, pow, sum, mean},
  breaklines=true,
  breakatwhitespace=true,
  columns=flexible,
  tabsize=2,
  showstringspaces=false,
}

\begin{algorithm}[htb]
\caption{PyTorch-style implementation of the MGM losses}
\label{alg:mgm_algo_code}
\begin{lstlisting}
# G:     generator network,        z -> x_0
# f:     critic network,           (x, t) -> symmetrized displacement
# sigma: interpolant noise level,  sigma_t = sqrt(sigma * t * (1 - t))

def flipped_interpolant(x_0, x_1, sigma):
  t     = rand(N, 1) * 0.5              # [N, 1], uniform in [0, 1/2]
  b     = (rand(N, 1) < 0.5).float()    # [N, 1], Bernoulli(1/2)
  noise = randn(N, D)                   # [N, D]

  # the two branches of the random time flip (sigma_t = sigma_{1-t})
  sigma_t = sqrt(sigma * t * (1 - t))
  x_t   = (1 - t) * x_0 + t * x_1 + sigma_t * noise
  x_1mt = t * x_0 + (1 - t) * x_1 + sigma_t * noise

  # hide which branch was observed
  x_tilde = (1 - b) * x_t + b * x_1mt   # [N, D]
  return x_tilde, t


def mgm_loss_f(G, f, sigma):
  # -------- critic: Flow Matching regression on x_tilde --------
  x_0 = stop_grad(G(sample_noise(N)))   # [N, D]
  x_1 = sample_data(N)                  # [N, D]
  x_tilde, t = flipped_interpolant(x_0, x_1, sigma)

  target = x_1 - x_0                    # [N, D]
  pred   = f(x_tilde, t)                # [N, D]
  return (pred - target).pow(2).sum(-1).mean()


def mgm_loss_G(G, f, sigma):
  # -------- generator: variational midpoint objective --------
  x_0 = G(sample_noise(N))              # [N, D]
  x_1 = sample_data(N)                  # [N, D]
  x_tilde, t = flipped_interpolant(x_0, x_1, sigma)

  target = x_1 - x_0                    # [N, D]
  pred   = f(x_tilde, t)                # [N, D]
  return (pred * (2 * target - pred)).sum(-1).mean()
\end{lstlisting}
\end{algorithm}

\section{Additional Experimental Details}\label{app:experimental_details}
In this section we provide details specifying the experimental setup of~\S\ref{sec:experiments}.
\subsection{General Details}\label{app:general_details}

\textbf{Implementation.}
Our implementation builds on the SiD codebase\footnote{\url{https://github.com/mingyuanzhou/SiD}}
\new{and is publicly available at} \url{https://github.com/daniil-shlenskii/mgm}.

\textbf{Data and architectures.}
On CIFAR-10 $32\times32$ and ImageNet $64\times64$, MGM operates in pixel space.
On CIFAR-10, the generator and the \critic\ use the DDPM++ architecture from EDM~\citep{karras2022elucidating} (56M parameters); on ImageNet $64\times64$, they use the ADM architecture~\citep{dhariwal2021diffusion} as trained in EDM (296M parameters).
On ImageNet $512\times512$, we follow the latent-space setup of EDM2~\citep{karras2024analyzing}: images are encoded by the Stable Diffusion VAE~\citep{rombach2022high} into $64\times64\times4$ latents, both networks use the EDM2-S architecture (280M parameters), and generated latents are decoded by the frozen VAE decoder before evaluation.
In all cases, the generator and the \critic\ share the same architecture and are initialized from the same pretrained network (Appendix~\ref{app:initialization}).

\textbf{Evaluation.}
We report FID computed on 50K generated samples against the reference statistics of the training set provided by EDM~\citep{karras2022elucidating} and EDM2~\citep{karras2024analyzing}, using the Inception network from the EDM evaluation code.
We evaluate EMA checkpoints of the generator and report the lowest FID obtained during training.

\textbf{Hyperparameters.}
Training hyperparameters are summarized in Table~\ref{tab:hyperparameters}. The generator and the \critic\ share the same optimizer settings. We maintain an exponential moving average (EMA) of the generator weights only, and the \critic\ is discarded after training.
We do not use data augmentation during MGM training.

\begin{table}[h]
    \centering
    \small
    \setlength{\tabcolsep}{4pt}
    \renewcommand{\arraystretch}{1.12}
    \begin{tabularx}{\textwidth}{l *{3}{>{\centering\arraybackslash}X}}
        \toprule
        & \textbf{CIFAR-10}
        & \textbf{ImageNet} $64\times64$
        & \textbf{ImageNet} $512\times512$ \\
        \midrule
        Space & pixel & pixel & latent (SD VAE) \\
        Architecture & DDPM++ (56M) & ADM (296M) & EDM2-S (280M) \\
        Initialization & 50K\new{~kimgs} warmup & EDM checkpoint & EDM2-S checkpoint \\
        \midrule
        Batch size & 256 & 2048 & 256 \\
        Adam $(\beta_1,\beta_2)$ & \multicolumn{3}{c}{$(0,0.999)$} \\
        Learning rate & \multicolumn{3}{c}{$10^{-5}$} \\
        Critic:generator updates & \multicolumn{3}{c}{1:1} \\
        Generator EMA decay & \multicolumn{3}{c}{0.999} \\
        \midrule
        Noise $\sigma$ & 0.01 & 0.2 & 0.5 \\
        Critic-only noise & \multicolumn{3}{c}{Yes} \\
        \bottomrule
    \end{tabularx}
    \vspace{5pt}
    \caption{Training hyperparameters used for the baseline comparison in \S\ref{subsec:baseline_comparison}.}
    \label{tab:hyperparameters}
\end{table}

\textbf{Resources.}
A single MGM training run takes approximately 2 days on 2 NVIDIA H100 GPUs for CIFAR-10, 3 days on 8 H100 GPUs for ImageNet $64\times64$, and 4 days on 8 H100 GPUs for ImageNet $512\times512$, all in full precision and excluding the initialization stage.

\subsection{Initialization and Preconditioning}\label{app:initialization}

\textbf{Pretrained denoiser.}
In all experiments, both networks \new{use the EDM-style denoiser parameterization}~\citep{karras2022elucidating}
\[
D_\eta(x,\sigma) = c_{\rm skip}(\sigma)\,x + c_{\rm out}(\sigma)\,F_\eta\big(c_{\rm in}(\sigma)\,x,\; c_{\rm noise}(\sigma)\big),
\]
\new{where $F_\eta$ is the raw network,} with the standard EDM preconditioning
$c_{\rm skip}(\sigma)=\sigma_{\rm data}^2/(\sigma^2+\sigma_{\rm data}^2)$,
$c_{\rm out}(\sigma)=\sigma\sigma_{\rm data}/\sqrt{\sigma^2+\sigma_{\rm data}^2}$,
$c_{\rm in}(\sigma)=1/\sqrt{\sigma^2+\sigma_{\rm data}^2}$,
$c_{\rm noise}(\sigma)=\tfrac14\log\sigma$\new{, and $\sigma_{\rm data}=0.5$}.
On CIFAR-10, $D_\eta$ is obtained by the short denoising warmup described in Appendix~\ref{app:warmup}; on ImageNet $64\times64$ and $512\times512$, we use the publicly available EDM and EDM2-S checkpoints, respectively. For class-conditional models, both networks additionally receive the class label.

\textbf{Generator.}
Following SiD~\citep{zhou2024score}, the generator is initialized as a one-step denoiser applied to scaled Gaussian noise:
\[
G_\theta(z) = D_\theta(\sigma_{\rm init} z,\, \sigma_{\rm init}), \qquad z\sim\mathcal N(0,\mathrm{Id}), \qquad \theta \leftarrow \eta,
\]
with $\sigma_{\rm init}=2.5$\new{, matching the maximal noise level used during warmup (Appendix~\ref{app:warmup})}.

\textbf{Critic.}
The \critic\ reuses the same network and preconditioning with two modifications:
\[
f_\psi(x,t) = D_\psi(x,\, \sigma_{\min}+t) - x, \qquad \psi \leftarrow \eta,
\]
\new{where $\sigma_{\min}=0.002$ matches the default minimum sampling noise level in EDM.}
First, since $c_{\rm noise}$ is undefined at $\sigma=0$, we shift the noise level by $\sigma_{\min}$\new{,} so that the \critic\ is well defined for all $t\in[0,1]$.
Second, the \critic\ outputs the residual $D_\psi-x$ \new{without time-dependent rescaling}. In Flow Matching, the velocity is obtained from the denoiser output by rescaling $D-x$ with a time-dependent factor. Under the random time flip, however, this factor depends on the unobserved flip $B$: it differs between the original and the flipped branch. To avoid committing to either branch, we omit the rescaling and let the network absorb the scale.

\subsection{Denoising Warmup}\label{app:warmup}

On CIFAR-10, we obtain the pretrained denoiser $D_\eta$ (Appendix~\ref{app:initialization}) by a short denoising warmup: we train the DDPM++ network from scratch with the EDM objective and training hyperparameters~\citep{karras2022elucidating} for 50K\new{~kimgs} (25\% of the EDM training budget), unless stated otherwise (Table~\ref{tab:ablation_warmup}). This warmup is used only to obtain a reasonable initialization; we do not aim to train a full diffusion model.

The only modification of EDM training concerns the sampled noise levels. EDM samples log-noise values
\[
n \sim \mathcal{N}(-1.2,\, 1.2^2), \qquad \sigma = \exp(n),
\]
which covers noise levels far beyond those used by MGM:
\new{the generator is evaluated at $\sigma_{\rm init}=2.5$, while the \critic\ uses conditioning noise levels in $[\sigma_{\min},\sigma_{\min}+1/2]$.}
We therefore restrict the noise levels to $\sigma\le\sigma_{\max}$ with $\sigma_{\max}=\sigma_{\rm init}=2.5$ by reflecting the log-noise value around $\log\sigma_{\max}$:
\[
s =
\begin{cases}
n, & n \le \log \sigma_{\max},\\
2\log \sigma_{\max} - n, & n > \log \sigma_{\max},
\end{cases}
\qquad
\sigma = \exp(s).
\]
\new{This folded log-normal distribution reallocates the high-noise tail to noise levels below $\sigma_{\max}$, concentrating the warmup budget on the range relevant for MGM.}

\section{Additional Results}\label{app:additional_results}

In this section we provide additional results extending experiments of~\S\ref{sec:experiments}.

\subsection{Trigonometric Interpolant}\label{app:trig}
To assess the sensitivity of MGM to the choice of the interpolation path, we replace the linear deterministic component $I_t(x_0,x_1)=(1-t)x_0+tx_1$ with the trigonometric interpolant~\citep{albergo2023building}
\[
I_t(x_0,x_1)=\cos\!\big(\tfrac{\pi t}{2}\big)\,x_0+\sin\!\big(\tfrac{\pi t}{2}\big)\,x_1 .
\]
It satisfies the endpoint conditions and the symmetry $I_t(x_0,x_1)=I_{1-t}(x_1,x_0)$ required in \S\ref{subsec:generalized_midpoint_divergence}, since $\cos(\pi(1-t)/2)=\sin(\pi t/2)$. We keep the noise schedule and all other settings of \new{the CIFAR-10 configuration used in the main comparison} fixed. The trigonometric interpolant reaches FID 2.52 compared to 2.27 for the linear one, indicating that MGM is not tied to a particular interpolation path.

\subsection{Critic-to-Generator Update Ratio}\label{app:update_ratio}
Since MGM is trained with a minimax objective, we examine its sensitivity to the number of \critic\ updates per generator update, keeping all other settings of \new{the CIFAR-10 configuration used in the main comparison} fixed. With update ratios 1:1, 2:1, and 4:1, MGM reaches FID 2.27, 1.99, and 3.21, respectively. The effect is non-monotonic: a moderate increase in \critic\ updates improves FID, whereas a larger one degrades it. Thus, MGM does not require multiple \critic\ updates per generator update, although tuning the ratio can further improve performance. \new{We use the 1:1 ratio throughout the main experiments for consistency.}

\subsection{Seed Variability}\label{app:seeds}
To quantify run-to-run variability, we repeat \new{the CIFAR-10 configuration used in the main comparison} with two additional random seeds. The three runs reach FID 2.27, 2.24, and 2.29, i.e., $2.27\pm0.02$ (mean $\pm$ standard deviation). This indicates low sensitivity to the random seed within the used training regime.

\subsection{Noise Level on ImageNet $64\times64$}\label{app:imagenet64_noise}
We repeat the noise ablations of \S\ref{subsec:ablation_study} on class-conditional ImageNet $64\times64$ (Table~\ref{tab:imagenet64_noise}). The results follow the same qualitative pattern as on CIFAR-10: the deterministic setting $\sigma=0$ performs poorly, moderate noise gives the best results in both settings, and large noise $\sigma=0.5$ degrades training, most severely when noise is added only to the \critic. Compared to CIFAR-10, however, the optimal noise level is larger: $\sigma=0.1$--$0.2$ instead of $0.01$--$0.05$. With critic-only noise at $\sigma=0.2$, MGM achieves FID 2.50, which we report in Table~\ref{tab:baseline_cifar10}.
\begin{table}[h]
    \centering
    \small
    \begin{tabular}{l c c c c c c}
        \toprule
        $\sigma$ & 0.00 & 0.01 & 0.05 & 0.10 & 0.20 & 0.50 \\
        \midrule
        Both interpolants & 5.40 & 4.90 & 2.67 & \textbf{2.54} & 2.74 & 15.71 \\
        Critic only       & 5.62 & 3.53 & 3.18 & 2.65 & \textbf{2.50} & $>100$ \\
        \bottomrule
    \end{tabular}
    \vspace{5pt}
    \caption{FID-50K on class-conditional ImageNet $64\times64$ for different noise levels $\sigma$, with noise added to both the generator and \critic\ interpolants or to the \critic\ interpolant only. The best result in each row is in \textbf{bold}.}
    \label{tab:imagenet64_noise}
\end{table}

\section{Electrostatic Interpretation for the Midpoint Divergence}\label{app:electrostatics}

Recently, electrostatics has inspired a new family of generative models~\citep{xu2022poisson,kolesov2025field,shlenskii2025overclocking}.
In this section, we show that the midpoint symmetry of Flow Matching established in~\S\ref{sec:method} has a natural connection to the method of images in electrostatics.
We first recall the electrostatic view of generative modeling.
We then turn to the method of images: a plane is equipotential whenever the charges on its two sides are mirror images of each other with opposite signs.
Finally, using the duality between flow and field matching~\citep{shlenskii2026unlocking}, we show how this mirror symmetry naturally suggests the vanishing midpoint velocity of Flow Matching.
This discussion is intended as intuition and is not used in our proofs.

\textbf{Electrostatic basics.}
EFM~\citep{kolesov2025field} augments the data space $\mathbb R^d$ with an extra coordinate $z$.
Points of $\mathbb R^{d+1}$ are written as $\tilde x=(x,z)$, and a vector field splits into its data and $z$-components, $\mathbf E=(E_x,E_z)$.
A unit point charge at $\tilde y\in\mathbb R^{d+1}$ creates the Coulomb field
\[
    \mathbf E_{\tilde y}(\tilde x)=\frac{1}{S_d}\,\frac{\tilde x-\tilde y}{\|\tilde x-\tilde y\|^{d+1}},
\]
where $S_d$ is the surface area of the unit sphere in $\mathbb R^{d+1}$.
\new{The Coulomb field is conservative, meaning that it can be written as the negative gradient of a scalar potential. For $d\ge2$, this potential is}
\[
    \new{\varphi_{\tilde y}(\tilde x)=\frac{1}{(d-1)S_d}\,\frac{1}{\|\tilde x-\tilde y\|^{d-1}},
    \qquad
    \mathbf E_{\tilde y}=-\nabla\varphi_{\tilde y}.}
\]
By the superposition principle, a charge density $q$ creates \new{the field and potential}
\[
    \mathbf E(\tilde x)=\int\mathbf E_{\tilde y}(\tilde x)\,q(\tilde y)\,d\tilde y,
    \qquad
    \new{\varphi(\tilde x)=\int\varphi_{\tilde y}(\tilde x)\,q(\tilde y)\,d\tilde y,}
\]
\new{with $\mathbf E=-\nabla\varphi$.}

\new{EFM represents two distributions as plates of a capacitor, placing $p_0$ as positive charge on the plane $z=0$ and $p_1$ as negative charge on the plane $z=L$. The resulting charge distribution is}
\[
    q(y,z)=p_0(y)\,\delta(z)-p_1(y)\,\delta(z-L).
\]
\new{Samples are transported between the plates along the field lines.
Inside the layer $0<z<L$, the positive plate repels and the negative plate attracts, so both contribute a positive $z$-component.
Hence $E_z>0$, and $z$ increases strictly along each field line and can serve as its parameter.
For a field line $(x(\tau),z(\tau))$ satisfying $d(x,z)/d\tau=\mathbf E(x,z)$, the chain rule gives}
\[
    \frac{dx}{dz}
    =\frac{dx/d\tau}{dz/d\tau}
    =\frac{E_x(x,z)}{E_z(x,z)}.
\]
Some field lines of the capacitor leave the layer, but only the field inside the layer matters for the discussion below.

\textbf{Method of images.}
\new{The classical method of images pairs charges with mirror charges of opposite signs.
Every point on the reflection plane is equally distant from the two charges in each pair, so their potential contributions cancel there.
When $p_0=p_1$, the EFM plates form exactly this configuration with respect to the midplane $z=L/2$.
The midplane is therefore equipotential, and the tangential field vanishes on it,}
\[
    \new{\varphi(x,L/2)=0,}
    \qquad
    E_x(x,L/2)=-\nabla_x\varphi(x,L/2)=0.
\]
\new{Since $E_z>0$, the field-line slope $dx/dz$ also vanishes at the midplane, providing the electrostatic counterpart of $v_{1/2}=0$.}

\new{The same mirror configuration gives $\varphi(x,L-z)=-\varphi(x,z)$ throughout the layer.
Differentiating in $x$ and $z$ yields}
\[
    E_x(x,L-z)=-E_x(x,z),
    \qquad
    E_z(x,L-z)=E_z(x,z)\new{.}
\]
\new{The field-line slope therefore satisfies}
\[
    \frac{E_x(x,L-z)}{E_z(x,L-z)}=-\frac{E_x(x,z)}{E_z(x,z)}.
\]

\textbf{From Field Matching to Flow Matching.}
\new{\citet{shlenskii2026unlocking} establish a duality between flow-based generative models and forward-only interaction fields in an augmented space.}
Under this duality, Flow Matching corresponds to an interaction field~\citep{manukhov2026interaction} on the layer $\mathbb R^d\times[0,1]$, where the time $t$ plays the role of the extra coordinate $z$ and $L=1$:
\[
    \mathbf E(x,t)=\big(v_t(x)\,p_t(x),\;p_t(x)\big),
    \qquad
    \frac{E_x(x,t)}{E_t(x,t)}=v_t(x)
    \quad\text{wherever } p_t(x)>0.
\]
This field is not a Coulomb field and, in general, not the gradient of any potential, so equipotentiality itself does not carry over to Flow Matching.
The slope of the field lines, however, plays the same role in both models: it is $E_x/E_z$ in EFM and the velocity $v_t$ in Flow Matching.
\new{For linear Flow Matching with independent endpoints and $p_0=p_1$, the analogous mirror symmetry is}
\[
    \new{p_{1-t}=p_t,}
    \qquad
    \new{v_{1-t}=-v_t,}
\]
\new{where the velocity antisymmetry was established by~\citet{wang2026zero}.}
\new{At $t=1/2$, this antisymmetry gives $v_{1/2}=0$, recovering Proposition~\ref{prop:midpoint_symmetry}.}

\section{Computational Cost}\label{app:computational_cost}
\begingroup
\new{

In this section, we compare the training cost of MGM with the 21 CIFAR-10 baselines in Table~\ref{tab:baseline_cifar10}.
A direct comparison is difficult: most papers omit accelerator count or wall-clock time, architectures differ in size, and training may require additional networks, initialization, trajectory sampling, or multiple stages.
Moreover, a single loss evaluation may involve several network calls, ODE integration, or Jacobian-vector products (JVPs).
Raw iteration and example counts are therefore not directly comparable.

\subsection{Training Cost Estimation}
We report compute in millions of forward-pass equivalents of a 56M-parameter network (\textbf{M FPE56}).
We use 56M because EDM DDPM++~\citep{karras2022elucidating} is the most common backbone in this CIFAR-10 comparison.
We assign the following costs:
\[
F=1,\qquad B=2,\qquad F_B=3,\qquad F_{\rm JVP}=2,\qquad F_{{\rm JVP},B}=4,
\]
where $F$ and $B$ denote a forward and a backward pass, $F_B$ a forward-backward pass, $F_{\rm JVP}$ a joint forward-JVP evaluation, and $F_{{\rm JVP},B}$ its evaluation with a backward pass.
For a network with $P$ million parameters, we scale the cost by $P/56$:
\begin{equation}\label{eq:fpe56}
\text{M FPE56}
=\frac{\text{iterations}\times\text{batch size}}{10^6}
\times\text{operation cost}\times\frac{P}{56}.
\end{equation}
We sum this expression over the required stages and network calls.
A frozen network still incurs a forward pass, and we count $F_B$ when gradients must propagate through it to its input.
Near-56M backbones are rounded to 56M.
We scale iCT-deep (112M), the StyleGAN2 generator and discriminator (19.95M and 20.72M), the DFNO student and teacher (65.77M and 60M), and the VCT encoder (1.6M) separately.

\textbf{Initialization and post-initialization cost.}
\emph{Post-init} is the work performed after the required checkpoints are available.
\emph{Init} is the work used to obtain these checkpoints, including teacher pretraining where applicable.
We report $\text{Total}=\text{Post-init}+\text{Init}$.
The standard 56M EDM initialization costs $200\times3=600$ M FPE56, corresponding to 200M processed training images.
Image counts include repeated presentations of the training data.
We include online teacher calls and offline trajectory generation when their budgets are available.
The estimates describe a single training pipeline and exclude hyperparameter searches, additional seeds, and FID evaluation.

\textbf{Scope of the estimates.}
FPE56 provides an approximate estimate of computational cost. It does not directly measure FLOPs, GPU-hours, or wall-clock time.
In particular, scaling by parameter count does not capture all differences in architectures, memory access, precision, or implementations.
When a budget is not reported, we mark it as unknown.
The notation $\geq$ denotes a lower bound under this accounting: the corresponding LPIPS, VGG, DSM, GAN, feature-loss, or R1 computations are not included.
These omissions are specified in Tables~\ref{tab:compute_details_no_teacher} and~\ref{tab:compute_details_teacher}.
Our measured training times are reported separately in Appendix~\ref{app:general_details}.

\subsection{MGM Training Cost}
We consider the main CIFAR-10 configuration with FID 2.27 and one \critic\ update per generator update.
The denoising warmup processes 50M images, one quarter of the 200M-image EDM initialization budget, and trains a single 56M network.
Its weights initialize both the generator and the \critic, so this warmup is counted once:
\[
C_{\rm init}=50\times3=150\quad\text{M FPE56}.
\]
The subsequent MGM stage uses $78{,}125$ iterations with batch size 256, corresponding to 20M examples for each update branch in Algorithm~\ref{alg:mgm_training_algo}.
The \critic\ update requires a stop-gradient generator forward and a \critic\ forward-backward pass, costing $F+F_B=4$ per example.
The generator update uses a fresh batch and requires a generator forward-backward pass and differentiation through the \critic\ to its input, costing $2F_B=6$.
Thus,
\[
C_{\rm post}=20\times(4+6)=200,\qquad
C_{\rm total}=150+200=350\quad\text{M FPE56}.
\]
This estimate includes both networks and their recomputation between updates.
It applies to the 1:1 configuration, rather than the 2:1 update-ratio ablation in Appendix~\ref{app:update_ratio}.

\subsection{Comparison with Baselines}
Table~\ref{tab:training_cost} summarizes the estimates, and Tables~\ref{tab:compute_details_no_teacher} and~\ref{tab:compute_details_teacher} give the schedules and per-operation calculations.
As in Table~\ref{tab:baseline_cifar10}, we distinguish whether a pretrained model provides a training signal, rather than whether initialization is used.
Within this comparison, MGM achieves the best FID without a teacher-model training signal and the third-best FID overall, behind CTM and SiD.
MGM has the lowest complete post-initialization estimate at 200 M FPE56.
The closest estimates are 204.8 for ECT, at least 204.8 for CMT w/ ECT, and 209.2 for VCT.
Including initialization gives 350 M FPE56 for MGM, compared with 804.8, at least 804.8, and 809.2 M FPE56, respectively.
Under this accounting, MGM reduces the total training cost by at least 56\% relative to these three alternatives.
This comparison uses the reported training configurations and does not establish a ranking at a matched training budget or a corresponding wall-clock speedup.
}

\begin{table}[!htbp]
\new{
\centering
\small
\setlength{\tabcolsep}{5pt}
\renewcommand{\arraystretch}{1.16}
\begin{tabular*}{\textwidth}{@{\extracolsep{\fill}}l r r r@{}}
\toprule
Method & Post-init & Init & Total \\
\midrule
\multicolumn{4}{c}{\textit{Without teacher-model training signal}} \\
\midrule
\textbf{MGM (ours)} & \textbf{200} & \textbf{150} & \textbf{350} \\
DiffRatio-DiJS~\citep{chen2025diffratio} & Unknown & 600 & Unknown ($\geq600$) \\
iCT-deep~\citep{song2024improved} & 3276.8 & 0 & 3276.8 \\
iCT~\citep{song2024improved} & 1638.4 & 0 & 1638.4 \\
sCT~\citep{lu2025simplifying} & Unknown & 614.4 & Unknown ($\geq614.4$) \\
MeanFlow~\citep{geng2025mean} & 3276.8 & 0 & 3276.8 \\
Stable CT~\citep{wang2025stable} & 614.4 & 600 & 1214.4 \\
StyleGAN2+ADA+tuning~\citep{karras2020training} & $\geq475.5$ & 0 & $\geq475.5$ \\
IMM~\citep{zhou2025inductive} & 6553.6 & 0 & 6553.6 \\
VCT~\citep{silvestri2025vct} & 209.2 & 600 & 809.2 \\
ECT~\citep{geng2025consistency} & 204.8 & 600 & 804.8 \\
\midrule
\multicolumn{4}{c}{\textit{With teacher-model training signal}} \\
\midrule
CTM~\citep{kim2024consistency} & $\geq802.1$ & 600 & $\geq1402.1$ \\
SiD~\citep{zhou2024score} & 6500 & 600 & 7100 \\
TCM~\citep{lee2025truncated} & 1331.2 & 600 & 1931.2 \\
CMT w/ ECT~\citep{hu2026cmt} & $\geq204.8$ & 600 & $\geq804.8$ \\
CD~\citep{song2023consistency} & $\geq2457.6$ & 614.4 & $\geq3072$ \\
sCD~\citep{lu2025simplifying} & Unknown & 614.4 & Unknown ($\geq614.4$) \\
DMD~\citep{yin2024one} & $\geq1311.1$ & 600 & $\geq1911.1$ \\
DFNO~\citep{zheng2023fast} & $\geq909.4$ & 329.1 & $\geq1238.5$ \\
TRACT~\citep{berthelot2023tract} & 1536 & 600 & 2136 \\
2-Rectified Flow~\citep{liu2022flow} & Unknown & Unknown & Unknown \\
PD~\citep{salimans2022progressive} & $[480,2048]$ & 614.4 & $[1094.4,2662.4]$ \\
\bottomrule
\end{tabular*}
\caption{\textbf{Training-cost estimates on unconditional CIFAR-10.}
All costs are in M FPE56. The MGM initialization is one quarter of the standard EDM initialization.
Unknown denotes an unreported budget, and $\geq$ indicates an incomplete cost estimate with the omissions listed in Tables~\ref{tab:compute_details_no_teacher} and~\ref{tab:compute_details_teacher}.
FID values are reported in Table~\ref{tab:baseline_cifar10}.}
\label{tab:training_cost}
}
\end{table}

\begin{table}[p]
\new{
\centering
\small
\setlength{\tabcolsep}{5pt}
\renewcommand{\arraystretch}{1.12}
\begin{tabularx}{\textwidth}{@{}p{0.19\textwidth} >{\raggedright\arraybackslash}X@{}}
\toprule
Method & Post-initialization schedule and calculation \\
\midrule
MGM & $78{,}125\times256=20$M examples per update branch.
One \critic\ update and one generator update cost $F+3F_B=10$ per example, giving $20\times10=200$. \\
\addlinespace
DiffRatio-DiJS & Batch size 64. The number of iterations and the checkpoint used for the reported FID are not specified, so post-init cost is unknown. \\
\addlinespace
iCT-deep & $400\mathrm{k}\times1024=409.6$M examples.
Each iteration uses $F+F_B=4$, with a 112M-parameter network:
$409.6\times4\times112/56=3276.8$. \\
\addlinespace
iCT & $400\mathrm{k}\times1024=409.6$M examples.
Each iteration uses $F+F_B=4$, giving $409.6\times4=1638.4$. \\
\addlinespace
sCT & Batch size 512, with $F_B+F_{\rm JVP}=5$ per example.
The number of training iterations is not reported.
Initialization uses 204.8M examples, giving $204.8\times3=614.4$. \\
\addlinespace
MeanFlow & $800\mathrm{k}\times1024=819.2$M examples.
Each iteration uses $F_{{\rm JVP},B}=4$, giving $819.2\times4=3276.8$. \\
\addlinespace
Stable CT & $300\mathrm{k}\times512=153.6$M examples.
Each iteration uses $F+F_B=4$, giving $153.6\times4=614.4$. \\
\addlinespace
StyleGAN2\newline +ADA+tuning & $1{,}562{,}500\times64=100$M examples.
The generator uses $F+F_B=4$ and the discriminator uses $3F_B=9$.
With 19.95M and 20.72M parameters, respectively, the estimate is
$\geq100\times(4\times19.95+9\times20.72)/56=475.5$, excluding R1. \\
\addlinespace
IMM & $400\mathrm{k}\times4096=1638.4$M examples.
Each iteration uses $F+F_B=4$, giving $1638.4\times4=6553.6$. \\
\addlinespace
VCT & $400\mathrm{k}\times128=51.2$M examples.
The backbone uses $F+F_B=4$, and the 1.6M-parameter encoder uses $F_B=3$:
$51.2\times(4\times56+3\times1.6)/56=209.2$. \\
\addlinespace
ECT & $400\mathrm{k}\times128=51.2$M examples.
Each iteration uses $F+F_B=4$, giving $51.2\times4=204.8$. \\
\bottomrule
\end{tabularx}
\caption{\textbf{Cost accounting without a teacher-model training signal.}
All computed costs are in M FPE56, $\mathrm{k}=10^3$, and M denotes $10^6$ processed examples.
Method references and initialization costs are given in Table~\ref{tab:training_cost}.}
\label{tab:compute_details_no_teacher}
}
\end{table}

\begin{table}[p]
\new{
\centering
\small
\setlength{\tabcolsep}{5pt}
\renewcommand{\arraystretch}{1.12}
\begin{tabularx}{\textwidth}{@{}p{0.19\textwidth} >{\raggedright\arraybackslash}X@{}}
\toprule
Method & Post-initialization schedule and calculation \\
\midrule
CTM & $100\mathrm{k}\times256=25.6$M examples.
Heun step counts $k\in\{1,\ldots,17\}$ are sampled proportionally to $k$, giving $2\sum k^2/\sum k\approx23.333$ teacher forwards per example.
The total is $25.333F+2F_B=31.333$, giving $\geq25.6\times31.333=802.1$, excluding DSM, GAN, and feature losses. \\
\addlinespace
SiD & $1{,}953{,}125\times256=500$M examples.
An update cycle uses $F+4F_B=13$, giving $500\times13=6500$. \\
\addlinespace
TCM & $250\mathrm{k}\times512+200\mathrm{k}\times1024=332.8$M examples across two stages.
Each stage uses $F+F_B=4$, giving $332.8\times4=1331.2$.
Stage 2 uses either the frozen Stage-1 or current Stage-2 target. \\
\addlinespace
CMT w/ ECT & $300\mathrm{k}\times128+100\mathrm{k}\times128=51.2$M examples.
Mid-training obtains 16 training pairs per 16-NFE teacher trajectory, averaging one teacher forward per pair.
ECT post-training instead uses one target forward per example.
Both stages additionally require a student forward-backward pass, giving $F+F_B=4$ per example and $\geq51.2\times4=204.8$ in total, excluding LPIPS. \\
\addlinespace
CD & $800\mathrm{k}\times512=409.6$M examples.
Each iteration uses $3F+F_B=6$, giving $\geq409.6\times6=2457.6$, excluding LPIPS. \\
\addlinespace
sCD & Batch size 512, with $F+F_{\rm JVP}+F_B=6$ per example.
The number of training iterations is not reported. \\
\addlinespace
DMD & $300\mathrm{k}\times392=117.6$M examples, with $2F+3F_B=11$ per example, plus 0.5M offline pairs at $35F$ per pair.
The estimate is $\geq117.6\times11+0.5\times35=1311.1$, excluding LPIPS. \\
\addlinespace
DFNO & $400\mathrm{k}\times256=102.4$M student examples with $F_B=3$, plus 1M teacher trajectories at $512F$ each.
Scaling the 65.77M student and 60M teacher separately gives
$\geq102.4\times3\times65.77/56+512\times60/56=909.4$, excluding VGG.
Teacher initialization costs 329.1. \\
\addlinespace
TRACT & $500\mathrm{k}\times512=256$M examples.
Each iteration uses $3F+F_B=6$, giving $256\times6=1536$. \\
\addlinespace
2-Rectified Flow & The reflow stage uses $300\mathrm{k}$ iterations, but its batch size, initial 1-RF training budget, pair-generation NFE, and final distillation budget are not reported. \\
\addlinespace
PD & In Consistency Models~\citep{song2023consistency}, the authors state: ``For progressive distillation, we used the same training settings as those described in Salimans \& Ho (2022).''
However, their \href{https://github.com/openai/consistency_models_cifar10/blob/d086e77dfdc30b51671685f9eec90c90d4f4eaa6/configs/cifar10_ve_progressive_distillation.py\#L15-L25}{released configuration} uses a batch size of 512 and approximately $800\mathrm{k}$ updates.
The original CIFAR-10 schedule uses a batch size of 128 and $750\mathrm{k}$ updates across all stages for one-step generation~\citep{salimans2022progressive}.
With $2F+F_B=5$ per example, these schedules give 480 and 2048 M FPE56, respectively.
To reflect this discrepancy, we report a post-initialization cost range of $[480,2048]$ M FPE56, or $[1094.4,2662.4]$ M FPE56 including initialization. \\
\bottomrule
\end{tabularx}
\caption{\textbf{Cost accounting with a teacher-model training signal.}
All computed costs are in M FPE56.
CD, sCD, and PD use a 204.8M-example initialization, costing 614.4 M FPE56.
The other initialization costs and method references are given in Table~\ref{tab:training_cost}.}
\label{tab:compute_details_teacher}
}
\end{table}
\clearpage
\endgroup

\clearpage
\section{Additional Samples from MGM}\label{app:additional_samples}
\vspace*{\fill}
\begin{figure}[H]
    \centering
    \includegraphics[width=\linewidth]{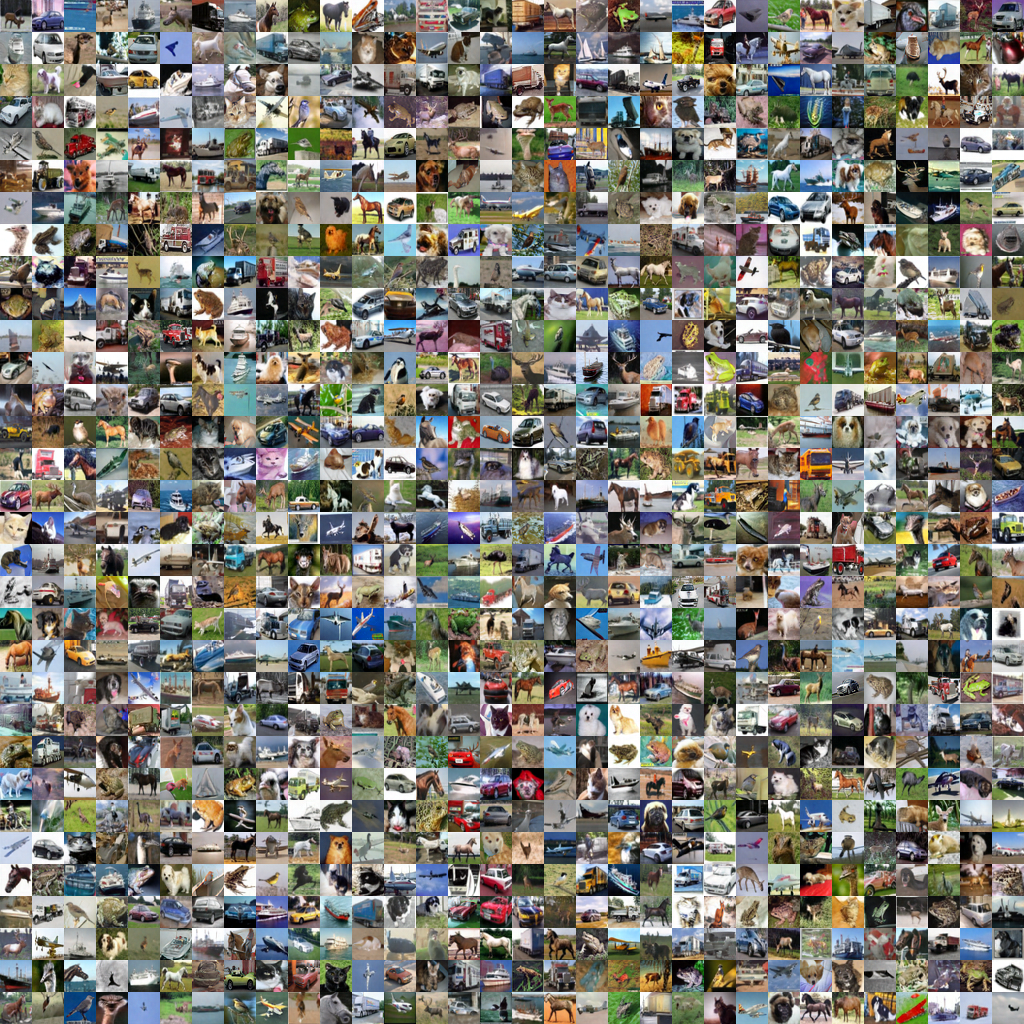}
    \caption{\new{One-step unconditional CIFAR-10 $32\times32$ samples generated by MGM (FID=2.27).}}
    \label{fig:cifar10_samples}
\end{figure}

\vspace*{\fill}
\clearpage
\vspace*{\fill}
\begin{figure}[H]
    \centering
    \includegraphics[width=\linewidth]{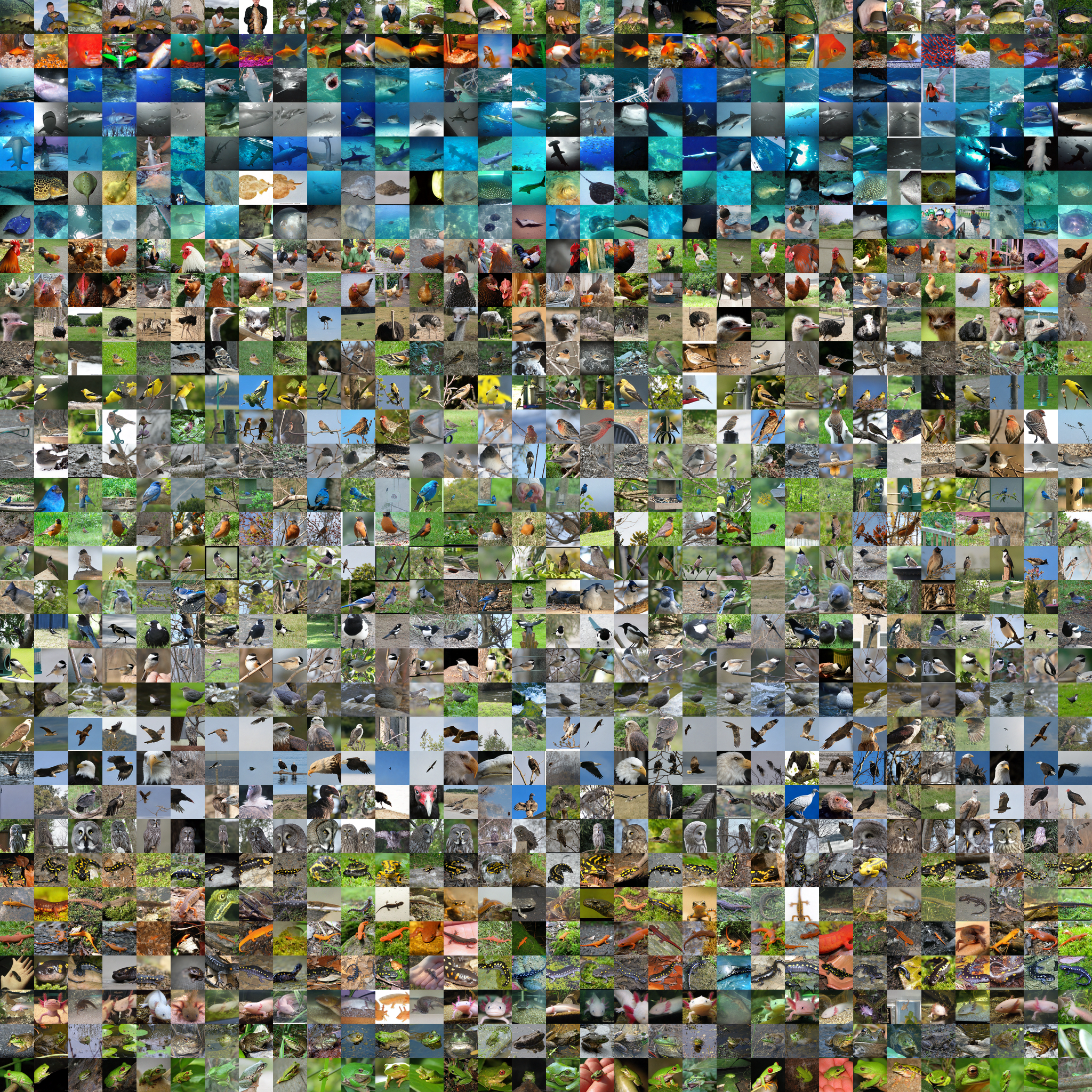}
    \caption{\new{One-step class-conditional ImageNet $64\times64$ samples generated by MGM (FID=2.50).}}
    \label{fig:imagenet64_samples}
\end{figure}

\vspace*{\fill}
\clearpage
\vspace*{\fill}
\begin{figure}[H]
    \centering
    \includegraphics[width=\linewidth]{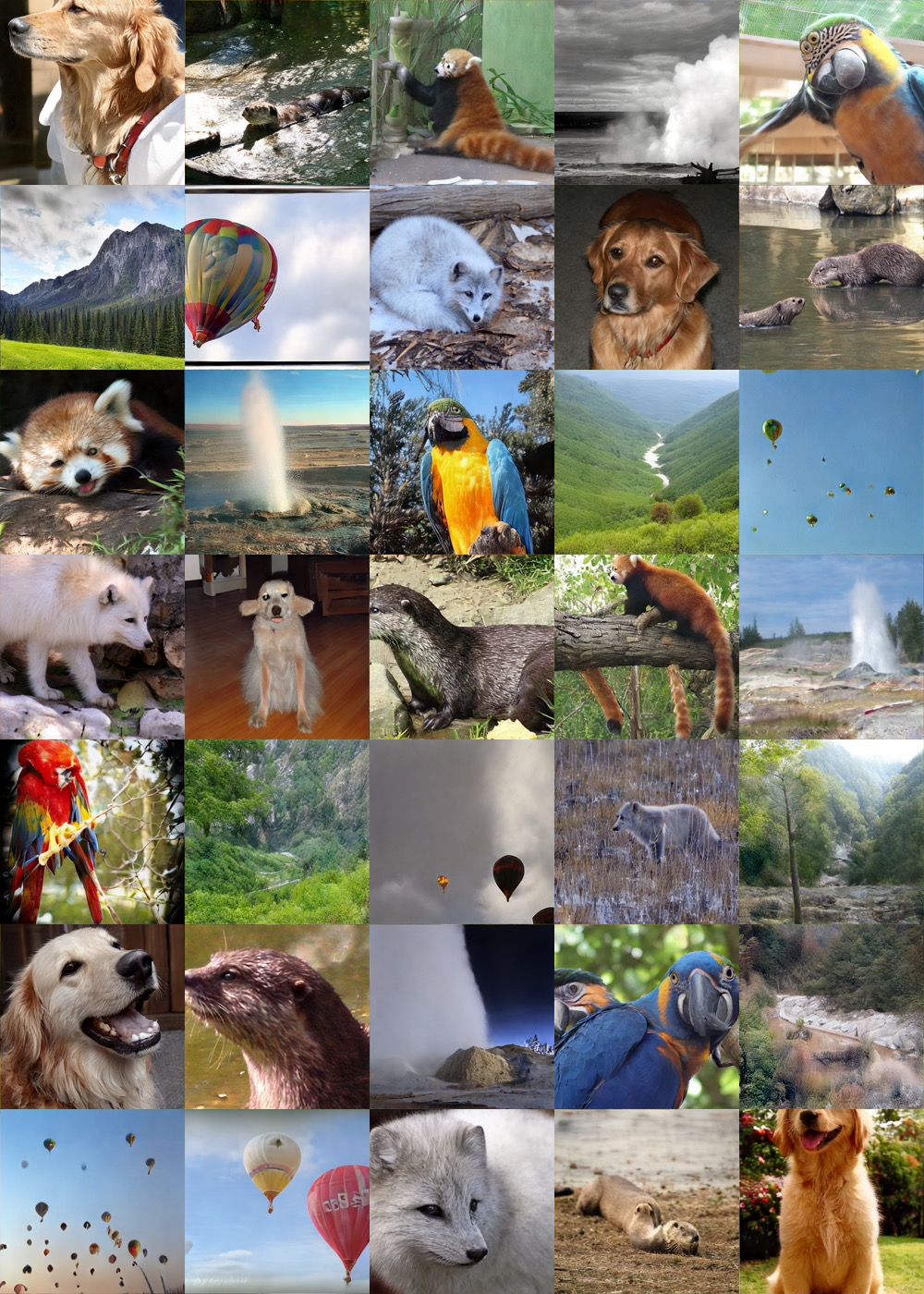}
    \caption{\new{One-step class-conditional ImageNet $512\times512$ samples generated by MGM (FID=2.36).}}
    \label{fig:imagenet512_appendix_samples}
\end{figure}

\vspace*{\fill}

\newpage
\section*{NeurIPS Paper Checklist}

\begin{enumerate}

\item {\bf Claims}
    \item[] Question: Do the main claims made in the abstract and introduction accurately reflect the paper's contributions and scope?
    \item[] Answer: \answerYes{} % Replace by \answerYes{}, \answerNo{}, or \answerNA{}.
    \item[] Justification: \new{The claims are supported by the theory in \S\ref{sec:method} and experiments in \S\ref{sec:experiments}, with limitations discussed in \S\ref{sec:discussion}.}

\item {\bf Limitations}
    \item[] Question: Does the paper discuss the limitations of the work performed by the authors?
    \item[] Answer: \answerYes{} % Replace by \answerYes{}, \answerNo{}, or \answerNA{}.
    \item[] Justification: \new{Limitations are discussed in \S\ref{sec:discussion}.}

\item {\bf Theory assumptions and proofs}
    \item[] Question: For each theoretical result, does the paper provide the full set of assumptions and a complete (and correct) proof?
    \item[] Answer: \answerYes{} % Replace by \answerYes{}, \answerNo{}, or \answerNA{}.
    \item[] Justification: \new{Assumptions and results are stated in \S\ref{sec:method}, with proofs in Appendix~\ref{app:appendix}.}

    \item {\bf Experimental result reproducibility}
    \item[] Question: Does the paper fully disclose all the information needed to reproduce the main experimental results of the paper to the extent that it affects the main claims and/or conclusions of the paper (regardless of whether the code and data are provided or not)?
    \item[] Answer: \new{\answerYes{}} % Replace by \answerYes{}, \answerNo{}, or \answerNA{}.
    \item[] Justification: \new{Algorithms and settings are given in Appendices~\ref{app:algorithm} and~\ref{app:experimental_details}.}

\item {\bf Open access to data and code}
    \item[] Question: Does the paper provide open access to the data and code, with sufficient instructions to faithfully reproduce the main experimental results, as described in supplemental material?
    \item[] Answer: \new{\answerYes{}} % Replace by \answerYes{}, \answerNo{}, or \answerNA{}.
    \item[] Justification: \new{Code and reproduction instructions are available in the repository linked in Appendix~\ref{app:general_details}.}

\item {\bf Experimental setting/details}
    \item[] Question: Does the paper specify all the training and test details (e.g., data splits, hyperparameters, how they were chosen, type of optimizer) necessary to understand the results?
    \item[] Answer: \answerYes{} % Replace by \answerYes{}, \answerNo{}, or \answerNA{}.
    \item[] Justification: \new{Experimental settings are described in \S\ref{sec:experiments} and Appendix~\ref{app:experimental_details}, with hyperparameters in Table~\ref{tab:hyperparameters}.}

\item {\bf Experiment statistical significance}
    \item[] Question: Does the paper report error bars suitably and correctly defined or other appropriate information about the statistical significance of the experiments?
    \item[] Answer: \new{\answerYes{}}% Replace by \answerYes{}, \answerNo{}, or \answerNA{}.
    \item[] Justification: \new{Appendix~\ref{app:seeds} reports the mean and standard deviation across three CIFAR-10 training seeds. Other experiments omit error bars due to computational cost.}

\item {\bf Experiments compute resources}
    \item[] Question: For each experiment, does the paper provide sufficient information on the computer resources (type of compute workers, memory, time of execution) needed to reproduce the experiments?
    \item[] Answer: \answerYes{} % Replace by \answerYes{}, \answerNo{}, or \answerNA{}.
    \item[] Justification: \new{Hardware and runtime per MGM run are reported in Appendix~\ref{app:general_details}.}
    
\item {\bf Code of ethics}
    \item[] Question: Does the research conducted in the paper conform, in every respect, with the NeurIPS Code of Ethics \url{https://neurips.cc/public/EthicsGuidelines}?
    \item[] Answer: \answerYes{} % Replace by \answerYes{}, \answerNo{}, or \answerNA{}.
    \item[] Justification: \new{The work conforms to the NeurIPS Code of Ethics.}

\item {\bf Broader impacts}
    \item[] Question: Does the paper discuss both potential positive societal impacts and negative societal impacts of the work performed?
    \item[] Answer: \answerNo{} % Replace by \answerYes{}, \answerNo{}, or \answerNA{}.
    \item[] Justification: \new{The paper focuses on methodology and benchmarks (\S\ref{sec:method} and \S\ref{sec:experiments}) and does not discuss societal impacts.}
    
\item {\bf Safeguards}
    \item[] Question: Does the paper describe safeguards that have been put in place for responsible release of data or models that have a high risk for misuse (e.g., pre-trained language models, image generators, or scraped datasets)?
    \item[] Answer: \answerNA{} % Replace by \answerYes{}, \answerNo{}, or \answerNA{}.
    \item[] Justification: \new{Experiments use standard image benchmarks (\S\ref{sec:experiments}); no high-risk assets are released.}

\item {\bf Licenses for existing assets}
    \item[] Question: Are the creators or original owners of assets (e.g., code, data, models), used in the paper, properly credited and are the license and terms of use explicitly mentioned and properly respected?
    \item[] Answer: \new{\answerNo{}} % Replace by \answerYes{}, \answerNo{}, or \answerNA{}.
    \item[] Justification: \new{Assets are credited in \S\ref{sec:experiments} and Appendix~\ref{app:general_details}, but licenses and terms are not explicitly listed.}

\item {\bf New assets}
    \item[] Question: Are new assets introduced in the paper well documented and is the documentation provided alongside the assets?
    \item[] Answer: \new{\answerYes{}} % Replace by \answerYes{}, \answerNo{}, or \answerNA{}.
    \item[] Justification: \new{The MGM implementation is documented in the repository linked in Appendix~\ref{app:general_details}.}

\item {\bf Crowdsourcing and research with human subjects}
    \item[] Question: For crowdsourcing experiments and research with human subjects, does the paper include the full text of instructions given to participants and screenshots, if applicable, as well as details about compensation (if any)? 
    \item[] Answer: \answerNA{} % Replace by \answerYes{}, \answerNo{}, or \answerNA{}.
    \item[] Justification: \new{No crowdsourcing or human-subject studies are involved.}

\item {\bf Institutional review board (IRB) approvals or equivalent for research with human subjects}
    \item[] Question: Does the paper describe potential risks incurred by study participants, whether such risks were disclosed to the subjects, and whether Institutional Review Board (IRB) approvals (or an equivalent approval/review based on the requirements of your country or institution) were obtained?
    \item[] Answer: \answerNA{} % Replace by \answerYes{}, \answerNo{}, or \answerNA{}.
    \item[] Justification: \new{No human-subject studies are involved.}

\item {\bf Declaration of LLM usage}
    \item[] Question: Does the paper describe the usage of LLMs if it is an important, original, or non-standard component of the core methods in this research? Note that if the LLM is used only for writing, editing, or formatting purposes and does \emph{not} impact the core methodology, scientific rigor, or originality of the research, declaration is not required.
    %this research? 
    \item[] Answer: \answerNA{} % Replace by \answerYes{}, \answerNo{}, or \answerNA{}.
    \item[] Justification: \new{LLMs were used only for writing, editing, and formatting.}

\end{enumerate}

% \section{Technical appendices and supplementary material}
% Technical appendices with additional results, figures, graphs, and proofs may be submitted with the paper submission before the full submission deadline (see above). You can upload a ZIP file for videos or code, but do not upload a separate PDF file for the appendix. There is no page limit for the technical appendices. 

% Note: Think of the appendix as ``optional reading'' for reviewers. The paper must be able to stand alone without the appendix; for example, adding critical experiments that support the main claims to an appendix is inappropriate. 

%%%%%%%%%%%%%%%%%%%%%%%%%%%%%%%%%%%%%%%%%%%%%%%%%%%%%%%%%%%%

\end{document}